\documentclass[10pt,twocolumn,letterpaper]{article}

\usepackage{iccv}
\usepackage{times}
\usepackage{epsfig}
\usepackage{graphicx}
\usepackage{subcaption} 
\usepackage{amsmath}
\usepackage{amsthm}
\usepackage{amssymb}
\usepackage{amsfonts}
\usepackage{mathabx}
\usepackage{multirow}

\usepackage[pagebackref=true,breaklinks=true,letterpaper=true,colorlinks,bookmarks=false]{hyperref}

\iccvfinalcopy 

\def\iccvPaperID{3875} 

\ificcvfinal\fi

\begin{document}

\newcommand{\methodname}{DONNA}
\newcommand{\fullmethodname}{Optimal Neural Architecture Distillation}

\title{Distilling Optimal Neural Networks: Rapid Search in Diverse Spaces}

\author{Bert Moons, Parham Noorzad, Andrii Skliar, Giovanni Mariani, \\ Dushyant Mehta, Chris Lott and Tijmen Blankevoort \\
Qualcomm AI Research\thanks{Qualcomm AI Research is an initiative of Qualcomm Technologies, Inc.}\\
{\tt\small \{bmoons,parham,askliar,gmariani,dushmeht,clott,tijmen\}@qti.qualcomm.com}
}

\maketitle
\ificcvfinal\thispagestyle{empty}\fi

\begin{abstract}
Current state-of-the-art Neural Architecture Search (NAS) methods neither efficiently scale to multiple hardware platforms, nor handle diverse architectural search-spaces. To remedy this, we present \methodname~(Distilling Optimal Neural Network Architectures), a novel pipeline for rapid, scalable and diverse NAS, that scales to many user scenarios. \methodname~consists of three phases. First, an accuracy predictor is built using blockwise knowledge distillation from a reference model. This predictor enables searching across diverse networks with varying macro-architectural parameters such as layer types and attention mechanisms, as well as across micro-architectural parameters such as block repeats and expansion rates. Second, a rapid evolutionary search finds a set of pareto-optimal architectures for any scenario using the accuracy predictor and on-device measurements. Third, optimal models are quickly finetuned to training-from-scratch accuracy. \methodname~ is up to 100$\times$ faster than MNasNet in finding state-of-the-art architectures on-device. Classifying ImageNet, \methodname~architectures are 20\% faster than EfficientNet-B0 and MobileNetV2 on a Nvidia V100 GPU and 10\% faster with 0.5\% higher accuracy than MobileNetV2-1.4x on a Samsung S20 smartphone. 
In addition to NAS, \methodname~is used for search-space extension and exploration, as well as hardware-aware model compression.
\end{abstract}

\section{Introduction} \label{sec:intro}
\begin{figure*}[ht]
\centering
\includegraphics[width=1.0\linewidth, trim=0 0 0 0, clip]{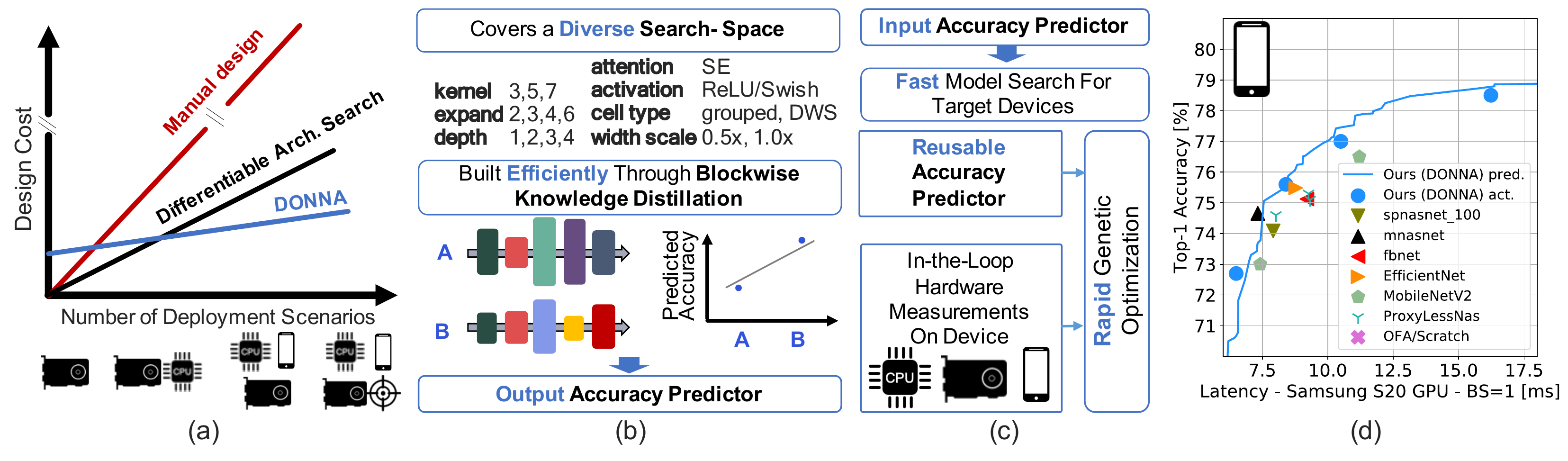}
\caption{Neural networks are deployed in many scenarios, on various hardware platforms with varying power modes and driver software, with different speed and accuracy requirements. \methodname~scales gracefully towards NAS for many of such scenarios, contrary to most prior approaches where NAS is repeated for each of them (a). This is achieved by splitting NAS into a scenario-agnostic training phase building an accuracy predictor through blockwise knowledge distillation (b) and a rapid scenario-aware search phase using this predictor and hardware measurements (c).
This yields a Pareto-front of models on-device, shown here for a Samsung S20 GPU on ImageNet~\cite{imagenet} (d).}
\label{fig:overview}
\vspace{-0.3cm}
\end{figure*}

Although convolutional neural networks (CNN) have achieved state-of-the-art performance for a wide range of vision tasks, they do not always execute efficiently on hardware platforms like desktop GPUs or mobile DSPs and NPUs.
To alleviate this issue, CNNs are specifically optimized to minimize latency and energy consumption for on-device performance. However, the optimal CNN architecture can vary significantly between different platforms. Even on a single platform, their efficiency can change with different operating conditions or driver versions. 
To solve this problem, low-cost methods for automated hardware-aware neural architecture search (NAS) are required. 

Current NAS algorithms, however, suffer from several limitations. First, many optimization algorithms~\cite{mnasnet,mbv3,single_path_nas,darts} target only a single \textit{deployment scenario}: a hardware-agnostic complexity metric, a hardware platform, or different latency, energy, or accuracy requirements. 
This means the search has to be repeated whenever any part of that scenario changes. Second, many methods cannot search in truly diverse search spaces, with different types of convolutional kernels, activation functions and attention mechanisms. Current methods either search through large and diverse spaces at a prohibitively expensive search cost~\cite{mnasnet,mbv3}, or limit their applicability by trading search time for a more constrained and less diverse search~\cite{ofa,single_path_nas,efficientnet,big_nas,nat,nsganetv2}. Most of such speedups in NAS come from a reliance on weight sharing mechanisms, which require all architectures in the search space to be structurally similar. Thus, these works typically only search among \textit{micro-architectural} choices such as kernel sizes, expansion rates, and block repeats and not among \textit{macro-architectural} choices of layer types, attention mechanisms and activation functions. As such, they rely on prior expensive methods such as~\cite{mnasnet,mbv3} for an optimal choice of macro-architecture. 

\begin{figure*}[h]
\centering
\includegraphics[width=.99\linewidth]{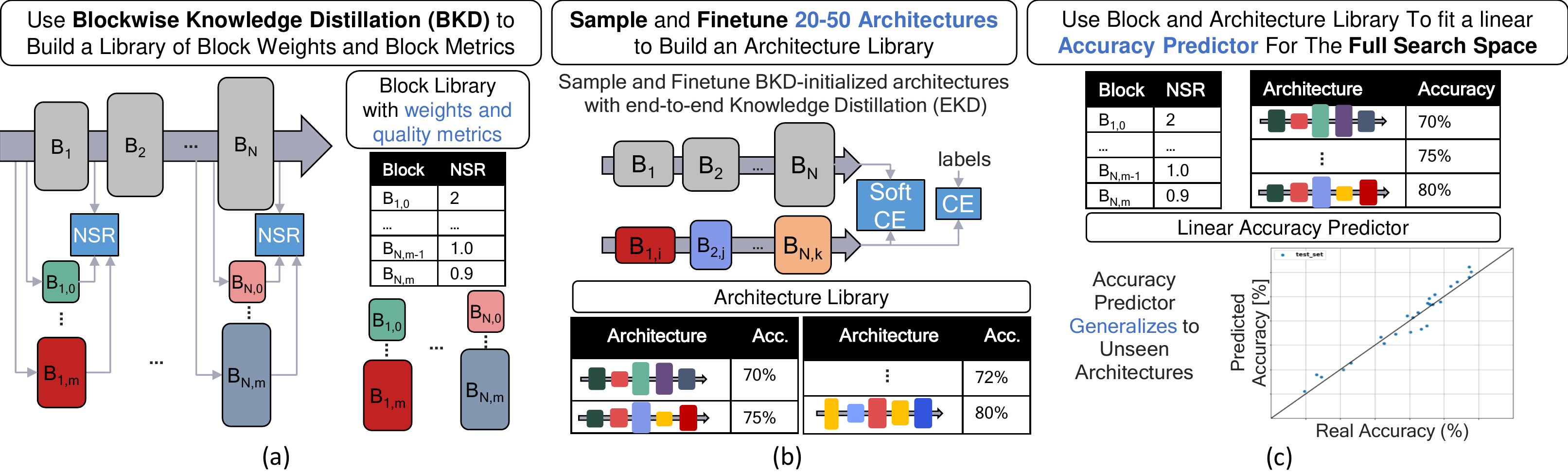}
\caption{An accuracy predictor is built in three steps. (a) Blockwise knowledge distillation (BKD) is executed to build a library of block-quality metrics and pretrained weights. (b) A set of full-model architectures is sampled from the search space and finetuned using the BKD initialization. (c) These results are used as targets to fit a linear accuracy predictor.}
\label{fig:bkd_overview}
\end{figure*}

We present \methodname~(Distilling Optimal Neural Network Architectures), a method that addresses both issues: it scales to multiple deployment scenarios with low additional cost, and performs rapid NAS in diverse search spaces. The method starts with a trained reference model. The first issue is resolved by splitting NAS into a scenario-agnostic training phase, and a scenario-aware search phase that requires only limited training, as depicted in Figure~\ref{fig:overview}. After an accuracy predictor is built in the training phase, the search is executed quickly for each new deployment scenario, typically in the time-frame of hours, and only requiring minimal fine-tuning to finalize optimal models. Second, \methodname~considers diverse \textit{macro-architectural} choices in addition to \textit{micro-architectural} choices, by creating this accuracy predictor through Blockwise Knowledge Distillation (BKD)~\cite{blockwise_nas}, see Figure~\ref{fig:bkd_overview}. This approach imposes little constraints on the macro- and micro-architectures under consideration, allowing a vast, diverse, and extensible search space. The \methodname~pipeline yields state of the art network architectures, as illustrated for a Samsung S20 GPU in Figure~\ref{fig:overview}(d). Finally, we use \methodname~for rapid search space extension and exploration, and on-device model compression. This is possible as the \methodname~accuracy predictor generalizes to architectures outside the original search space. 

\begin{figure*}[h]
\centering
\includegraphics[width=0.95\linewidth]{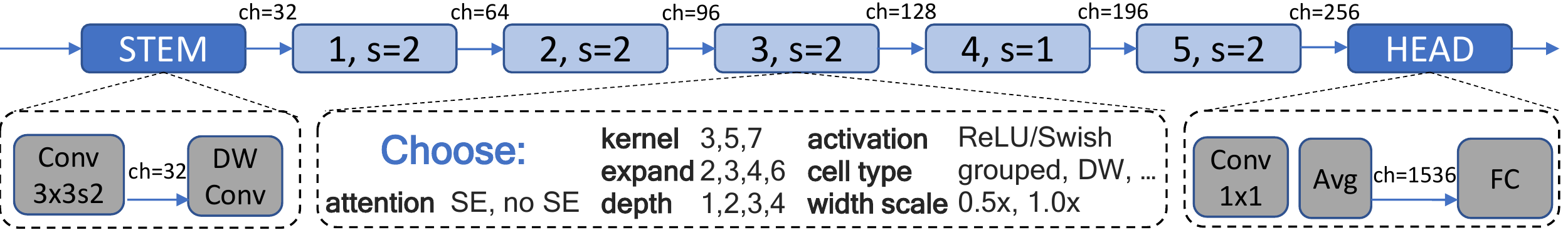}
\caption{\methodname~splits a model in a stem, head and N blocks. The search space is defined over the N blocks with varying kernel size, expand, depth, activation, cell type, attention and width scale factors. Block strides are kept constant.}
\label{fig:quarts_search_space}
\vspace{-0.3cm}
\end{figure*}
\section{Related Work} \label{sec:related_work}


Over time, methods in the NAS literature have evolved from prohibitively expensive but holistic and diverse search methods \cite{nas_rl, nasnet, mnasnet} to lower cost approaches that search in more restrictive non-diverse search spaces \cite{ofa,single_path_nas}. This work, \methodname, aims at benefiting from the best of both worlds: rapid search in diverse spaces. We refer the interested reader to the existing dedicated survey of Elsken et al. \cite{elsken2019neural} for a broader discussion of the NAS literature. 

Early approaches to NAS rely on reinforcement learning \cite{nas_rl, nasnet, mnasnet} or evolutionary optimization \cite{evolution_nas}. These methods allow for diverse search spaces, but at infeasibly high costs due to the requirement to train thousands of models for a number of epochs throughout the search. MNasNet \cite{mnasnet} for example uses up to 40,000 epochs in a single search. This process can be sped up by using weight sharing among different models, as in ENAS \cite{enas}. However, this comes at the cost of a less diverse search space, as the subsampled models have to be similar for the weights to be shareable.

In another line of work, differentiable architecture search methods such as DARTS \cite{darts}, FBNet~\cite{fbnet}, FBNetV2~\cite{fbnetv2}, ProxylessNAS \cite{proxylessnas}, AtomNAS \cite{atomnas} and Single-Path NAS \cite{single_path_nas} simultaneously optimize the weights of a large supernet and its architectural parameters. 
This poses several impediments to scalable and scenario-aware NAS in diverse search spaces. First, in most of these works, different cell choices have to be available to the algorithm, ultimately limiting the  space's size and diversity. While several works address this problem either by trading off the number of architecture parameters against the number of weights that are in GPU memory at a given time \cite{p_darts}, by updating only a subset of the weights during the search \cite{pc_darts}, or by exploiting more granular forms of weight-sharing~\cite{single_path_nas}, the fundamental problem remains when new operations are introduced. Second, although differentiable search methods speed up a single search iteration, the search must be repeated for every scenario due to their coupling of accuracy and complexity. Differentiable methods also require differentiable cost models. 
Typically these models use the sum of layer latencies as a proxy for the network latency, which can be inaccurate. This is especially the case in emerging depth-first processors \cite{depth_first}, where intermediate results are stored in the local memory, making full-graph latency depend on layer sequences rather than on individual layers. 


To improve the scaling performance of NAS across different scenarios, it is critical to decouple the accuracy prediction of a model from the complexity objective. In Once-for-All (OFA) \cite{ofa} and~\cite{nsganetv2}, a large weight-sharing supernet is trained using progressive shrinking. This process allows the sampling of smaller subnets from the trained supernet that perform comparably with models that have been trained from scratch. A large number of networks can then be sampled to build an accuracy predictor for this search space, which in turn can be used in a scenario-aware evolutionary search, as in Figure~\ref{fig:overview}(c). 
Although similar to \methodname~in this approach, OFA \cite{ofa} has several disadvantages. First, its search space's diversity is limited due to its reliance on progressive shrinking and weight sharing, which requires a fixed macro-architecture in terms of layer types, attention, activations, and channel widths.
Furthermore, progressive shrinking can only be parallelized in the batch dimension, limiting the maximum number of GPUs that can process in parallel. \methodname~does not suffer from these constraints.

Similarly, Blockwisely-Supervised NAS (DNA) \cite{blockwise_nas}, splits NAS into two phases: the creation of a ranking model for a search space and a targeted search to find the highest-ranked models at a given constraint. To build this ranking model, DNA uses blockwise knowledge distillation (BKD) to build a relative ranking of all possible networks in a given search space. The best networks are then trained and verified. It is crucial to note that it is BKD that 
enables the diverse search for optimal attention mechanisms, activation functions, and channel scaling. However, DNA has three disadvantages: (1) the ranking model fails when ranking large and diverse search spaces (Section~\ref{subsec:search_method/accuracy_model}), (2) the ranking only holds within a search space and does not allow the comparison of different spaces, and (3) because of the reliance on training subsampled architectures from scratch, the method is not competitive in terms of search time. This work, \methodname, addresses all these issues. In summary, DONNA differs from prior work on these key aspects:

\begin{enumerate}
    \item Unlike OFA \cite{ofa}, \methodname~enables hardware-aware search in \emph{diverse search spaces;} differentiable and RL-/evolutionary-based methods can do this too, but using much more memory or training time, respectively. 
    
    \item \methodname~\emph{scales to multiple accuracy/latency targets}, requiring only marginal cost for every new target. This is in contrast with differentiable or RL-/evolutionary-based methods, where the search has to be repeated for every new target.
    
    \item DONNA uses \emph{a novel accuracy predictor} which correlates better with training-from-scratch accuracy than prior work like DNA \cite{blockwise_nas} (See Figure \ref{fig:linear_predictor}).
    
    \item Furthermore, the DONNA accuracy predictor \emph{generalizes to unseen search spaces} due to its reliance on block \textit{quality metrics}, not on the network configuration (See Figure \ref{fig:search_space_analysis}).
    
    \item DONNA relies on a \emph{fast finetuning} method that achieves the same accuracy as training-from-scratch while being $9\times$ faster, reducing the training time for found architectures compared to DNA \cite{blockwise_nas}.
\end{enumerate}

\section{Distilling Optimal Neural Networks} \label{sec:search_method}
Starting with a trained reference model, \methodname~is a three step pipeline for NAS. For a given search space (Section~\ref{subsec:search_method/search-space}), we first build a scenario-agnostic accuracy predictor using Blockwise Knowledge Distillation (BKD) (Section~\ref{subsec:search_method/accuracy_model}). This amounts to a one-time cost. Second, a rapid scenario-aware evolutionary search phase finds the Pareto-optimal network architectures for any specific scenario (Section~\ref{subsec:search_method/evolutionary_search}). Third, the predicted Pareto-optimal architectures can be quickly finetuned up to full accuracy for deployment (Section~\ref{subsubsec:search_method/finetuning}).



\subsection{Search Space Structure}
\label{subsec:search_method/search-space}
Figure~\ref{fig:quarts_search_space} illustrates the block-level architecture of our search spaces and some parameters that can be varied within it. This search space is comprised of a stem, head, and $N$ variable blocks, each with a fixed stride. The choice of stem, head and the stride pattern depends on the choice of the reference model. The blocks used here are comprised of repeated layers, linked together by feedforward and residual connections. 
The blocks in the search space are denoted $B_{n,m}$, where $B_{n,m}$ is the $m^{th}$ potential replacement out of $M$ choices for block $B_n$ in the reference model. These blocks can be of any style of neural architecture (See Appendix~\ref{sec:supplementary/vit} for Vision Transformers~\cite{dosovitskiy2020image}), with very few structural limitations; only the spatial dimensions of the input and output tensors of $B_{n,m}$ need to match those of the reference model, which allows for diverse search. Throughout the text and in Appendix~\ref{sec:supplementary}, other reference models based on MobileNetV3~\cite{mbv3} and EfficientNet~\cite{efficientnet} are discussed.

\subsection{Building a Model Accuracy Predictor}
\label{subsec:search_method/accuracy_model}
\subsubsection{Blockwise Knowledge Distillation}
\label{subsubsec:search_method/bkd}

We discuss Blockwise Knowledge Distillation (BKD) as the first step in building an accuracy predictor for our search space, see Figure~\ref{fig:bkd_overview}(a). 
BKD yields a \textit{Block Library} of pretrained weights and quality metrics for each of the replacement blocks $B_{n,m}$. This is later used for fast finetuning (Section~\ref{subsubsec:search_method/finetuning}) and to fit the accuracy predictor (Section~\ref{subsec:search_method/linear_accuracy_predictor}). To build this library, each block $B_{n,m}$ is trained independently as a student using the pretrained reference block $B_{n}$ as a teacher. The errors between the teacher's output feature map $Y_n$ and the student's output feature map $\bar{Y}_{n,m}$ are used in this process. Formally, this is done by
minimizing the per-channel noise-to-signal-power ratio (NSR): 
\begin{equation}
\label{eq:loss}
\mathcal{L}(W_{n,m};Y_{n-1},Y_n) = \frac{1}{C}\sum_{c=0}^{C}
\frac{\|Y_{n,c}-\bar{Y}_{n,m,c}\|^2}{\sigma_{n,c}^2}
\end{equation}
Here, $C$ is the number of channels in a feature map,
$W_{n,m}$ are the weights of block $B_{n,m}$, $Y_n$ is the target output feature map of $B_n$, $Y'_{n,m}$ is the output of block $B_{n,m}$ and $\sigma_{n,c}^2$ is the variance of $Y_{n,c}$. This metric is closely related to Mean-Square-Error (MSE) on the feature maps, which~\cite{adaround} shows to be correlated to the task loss.

Essentially, the blocks $B_{n,m}$ are trained to closely replicate the teacher's non-linear function $Y_n = B_n(Y_{n-1})$. Intuitively, larger, more accurate blocks with a larger ``modeling capacity'' or ``expressivity'' replicate this function more closely than smaller, less accurate blocks. On ImageNet~\cite{imagenet} such knowledge distillation requires only a \emph{single} epoch of training for effective results. After training each block, the resulting NSR metric is added to the Block library as a \textit{quality metric} of the block $B_{n,m}$. Note that the total number of trainable blocks $B_{n,m}$ grows linearly as $N\times M$, whereas the overall search space grows exponentially as $M^N$, making the method scale well even for large search-spaces. 



\begin{figure}
\centering
\includegraphics[width=.99\linewidth]{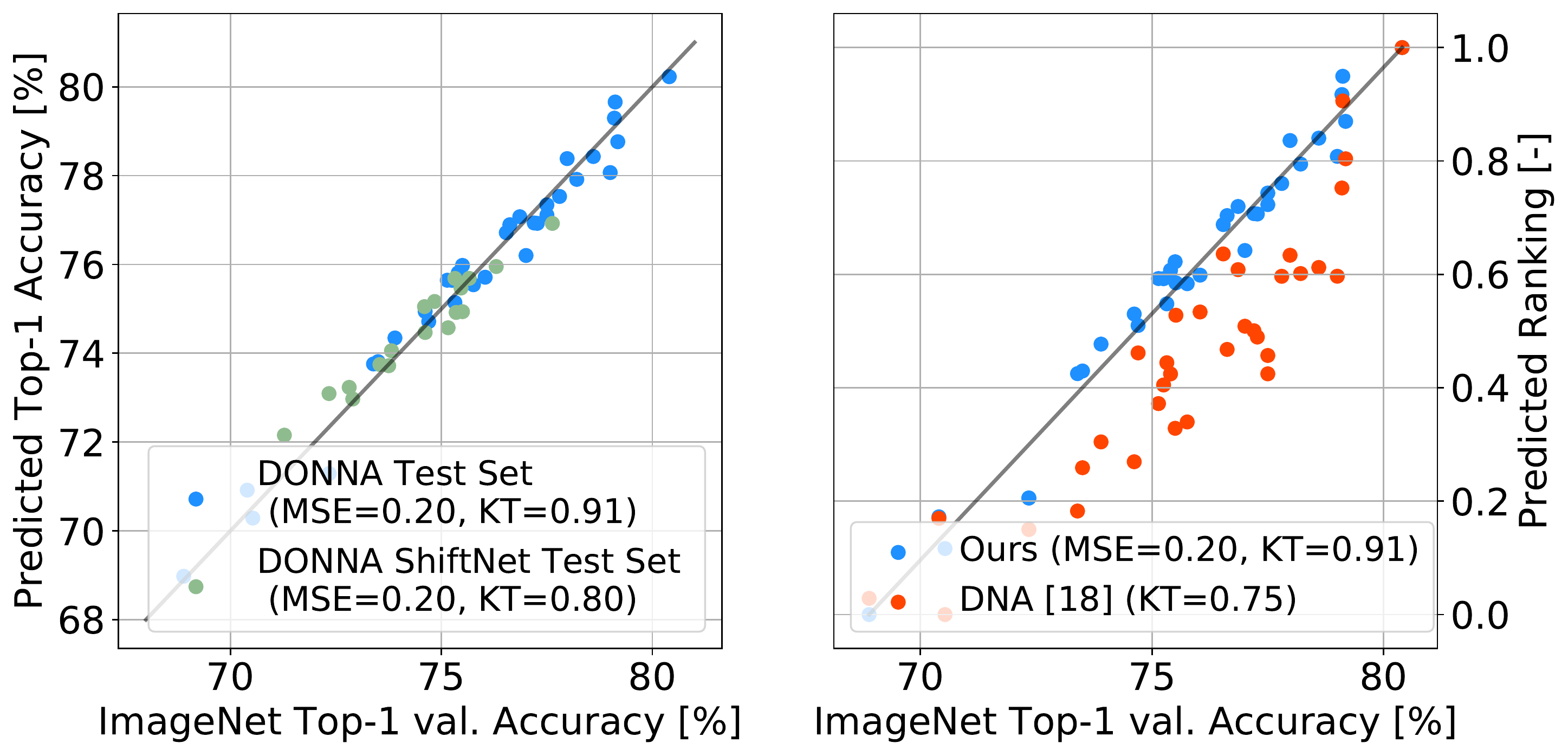}
\caption{The linear accuracy predictor generalizes to a test-set of unseen models (left), and is a better ranking predictor than DNA~\cite{blockwise_nas} (right) on the same set: Kendall-Tau~\cite{kendall1938new} of 0.91 in this work versus 0.75 for DNA.}
\label{fig:linear_predictor}
\end{figure}

\subsubsection{Linear Accuracy Predictor}
\label{subsec:search_method/linear_accuracy_predictor}


The key insight behind \methodname~is that block-level quality metrics derived through BKD (e.g., per-block NSR) can be used to predict the accuracy of all architectures sampled from the search space. We later show this metric even works for architectures outside of the search space (Section~\ref{subsubsec:experiments/search-space-design}).

To create an accuracy predictor, we build an \textit{Architecture Library} of trained models sampled from the search space, see Figure~\ref{fig:bkd_overview}(b). These models can be trained from scratch or finetuned quickly using weight initialization from BKD (Section~\ref{subsubsec:search_method/finetuning}).
Subsequently, we fit a linear regression model, typically using second-order terms, to predict the full search space's accuracy using the quality metrics stored in the Block Library as features and the accuracy from the Architecture Library as targets. Figure~\ref{fig:linear_predictor}(left) shows that the linear predictor fits well with a test-set of network architectures trained on ImageNet~\cite{imagenet} in the~\methodname~space (MSE=0.2, KT~\cite{kendall1938new}=0.91). This predictor can be understood as a sensitivity model that indicates which blocks should be large, and which ones can be small, to build networks with high accuracy. Appendix~\ref{subsubsec:supplementary/comparing_quality_metrics} discusses the effectiveness of different derived quality metrics on the quality of the accuracy prediction.


This process is now compared to DNA~\cite{blockwise_nas}, where BKD is used to build a ranking-model rather than an accuracy model. DNA~\cite{blockwise_nas} ranks subsampled architectures $i$ as:
\begin{equation}
\label{eq:dna_ranking}
R_i = \sum_{n=0}^{N}\frac{\|Y_{n}-\bar{Y}_{n,m_i}\|_1}{\sigma_{n}}
\end{equation}
which is sub-optimal due to two reasons. First, a ranking model only ranks models within the same search space and does not allow comparing performance of different search spaces. Second, the simple sum of quality metrics does not take the potentially different noise-sensitivity of blocks into account, for which a weighted sensitivity model is required. The \methodname~predictor takes on both roles.
Figure~\ref{fig:linear_predictor}(right) illustrates the performance of the linear predictor for the~\methodname~search space and compares the quality of its ranking to DNA~\cite{blockwise_nas}. Note that the quality of the \methodname~predictor increases over time, as whenever Pareto-optimal networks are finetuned, they can be added to the Architecture Library, and the predictor can be fitted again. 

\subsection{Evolutionary Search}
\label{subsec:search_method/evolutionary_search}
Given the accuracy model and the block library, the NSGA-II \cite{nsga_ii, pymoo} evolutionary algorithm is executed to find Pareto-optimal architectures that maximize model accuracy and minimize a target cost function, see Figure~\ref{fig:overview}(c). The cost function can be scenario-agnostic, such as the number of operations or the number of parameters in the network, or scenario-aware, such as on-device latency, throughput, or energy. In this work, full-network latency is considered as a cost function by using direct hardware measurements in the optimization loop. At the end of this process, the Pareto-optimal models yielded by the NSGA-II are finetuned to obtain the final models (Section~\ref{subsubsec:search_method/finetuning}). 

\subsection{Finetuning Architectures}
\label{subsubsec:search_method/finetuning}
Full architectures sampled from the search space can be quickly finetuned to match the from-scratch training accuracy by initializing them with weights from the BKD process (Section~\ref{subsubsec:search_method/bkd}). Finetuning is further sped up by using end-to-end knowledge distillation (EKD) using the reference model as a teacher, see Figure~\ref{fig:bkd_overview}(b). 
In Appendix~\ref{subsec:supplementary/finetuning-speed}, we show such models can be finetuned up to state-of-the-art accuracy in less than 50 epochs. This is a $9\times$ speedup compared to the state-of-the-art 450 epochs required in \cite{rw_imagenet} for training EfficientNet-style networks from scratch. 
This rapid training scheme is crucial to the overall efficiency of \methodname, since we use it for both, generating training targets for the linear accuracy predictor in Section \ref{subsec:search_method/accuracy_model}, as well as to finetune and verify Pareto-optimal architectures. 

\section{Experiments} \label{sec:experiments}

This section discusses three use-cases of \methodname: scenario-aware neural architecture search (Section~\ref{subsec:experiments/nas-for-imagenet}), search-space extrapolation and design (Section~\ref{subsubsec:experiments/search-space-design}), and model compression (Section~\ref{subsubsec:experiments/compression}). We also show that DONNA can be directly applied to object detection on MS-COCO~\cite{coco} and that architectures found by \methodname~transfer to optimal detection backbones (Section~\ref{subsec:experiments/detection}). DONNA is compared to random search in Appendix~\ref{sec:supplementary/random-search}.

\subsection{ImageNet Classification}
\label{subsec:experiments/imagenet}

\begin{figure*}[!t]
\centering
 \includegraphics[width=.99\linewidth]{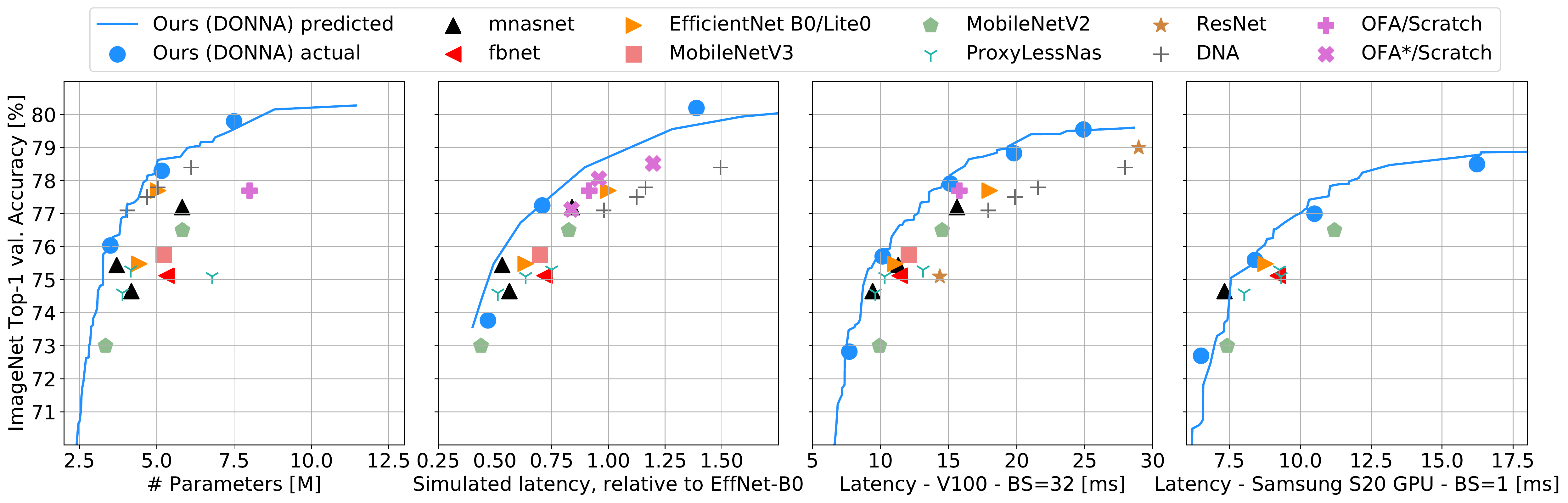}
\caption{The predicted Pareto-optimal front and models found by~\methodname~in the~\methodname~search space. Results are shown targeting the number of operations (left), the number of parameters (mid left), latency on a Nvidia V100 GPU (mid right) and latency on a simulator targeting tensor compute units in a mobile SoC (right). The trend line indicates predicted accuracy, whereas the dots are sampled from the trend line and finetuned up to the level of from-scratch accuracy. 
OFA*/Scratch results are our own search results using the framework in~\cite{ofa_repo} for $224\times224$ images, where the best models are retrained from scratch with DONNA hyperparameters for fair comparison.}

\label{fig:trendlines_overview}
\end{figure*}

\begin{table*}[h]
\caption{\label{tab:comparison}Comparing the cost of NAS methods, assuming 10 trained architectures per deployment scenario.
\methodname~can search in a diverse space similar to MNasNet~\cite{mnasnet} at a $100\times$ lower search-cost.}
\centering
\bgroup
\def\arraystretch{1.0}%
\begin{tabular}{c|c|c|c|c|c}
\multirow{2}{*}{Method} & \multirow{2}{*}{Granularity} & \multirow{2}{*}{Macro-Diversity} & Search-cost &  Cost / Scenario & Cost / Scenario \\
& & & 1 scenario [epochs] &  4 scenarios [epochs] &  $\infty$ scenarios [epochs] \\
\hline
OFA~\cite{ofa}& layer-level & fixed & 1200$+$10$\times [25 - 75]$ & $550-1050$ & $250-750$ \\
NSGANetV2~\cite{nsganetv2} & layer-level & fixed & 1200$+$10$\times [25 - 75]$ & $550-1050$ & $250-750$ \\
DNA~\cite{blockwise_nas} & layer-level & fixed & 770$+$10$\times$450 & $4700$ & $4500$ \\
MNasNet~\cite{mnasnet} & block-level & variable & 40000$+$10$\times$450 & $44500$ & $44500$ \\ \hline
This work & block-level & variable & 4000 $+$ 10$\times$50 &  $1500$ & $500$ \\
\end{tabular}
\egroup
\end{table*}
We present experiments for different search spaces for ImageNet classification: \methodname, EfficientNet-Compression and MobileNetV3 (1.0$\times$, 1.2$\times$). The latter two search spaces are \emph{blockwise} versions of the spaces considered by OFA \cite{ofa_repo}; that is, parameters such as expansion ratio and kernel size are modified on the block level rather than the layer level, rendering the overall search space coarser than that of OFA. Selected results for these spaces are discussed in this section, more extensive results can be found in Appendix~\ref{subsec:supplementary/various-search-spaces}. We first show that networks found by \methodname~in the \methodname~search space outperform the state-of-the-art (Figure~\ref{fig:trendlines_overview}). For example, \methodname~is up to $2.4\%$ more accurate on ImageNet~\cite{imagenet} validation compared to OFA\cite{ofa} trained from scratch with the same amount of parameters. At the same time, \methodname~finds models outperforming DNA \cite{blockwise_nas} up to 1.5$\%$ on a V100 GPU at the same latency and MobileNetV2 ($1.4\times$) by 10$\%$ at 0.5$\%$ higher accuracy on the Samsung S20 GPU. We also show that MobileNetV3-style networks found by \methodname~achieve the same quality of models compared to Mnasnet~\cite{mnasnet} and OFA~\cite{ofa} when optimizing for the same metric (See Fig.~\ref{fig:donna-finds-ofa-models} and Tab.~\ref{tab:mobilenetv3_comparison}). All experiments are for ImageNet~\cite{imagenet} images with $224\times224$ input resolution. Training hyperparameters are discussed in Appendix~\ref{subsec:supplementary/training_hyperparameters}.

\subsubsection{NAS for~\methodname~on ImageNet}
\label{subsec:experiments/nas-for-imagenet}
\methodname~is used for \textit{scenario-aware Neural Architecture Search} on ImageNet~\cite{imagenet}, quickly finding state-of-the-art models for a variety of deployment scenarios, see Figure~\ref{fig:trendlines_overview}. 

As shown in Figure~\ref{fig:quarts_search_space}, all 5 blocks $B_n$  in the \methodname~space can be replaced by a choice out of $M=384$ options: k $\in$ \{3,5,7\}; expand $\in$ \{2,3,4,6\}; depth $\in$ \{1,2,3,4\}; activation/attention $\in$ \{ReLU/None, Swish\cite{mbv3}/SE\cite{squeeze_excite}\}; layer-type $\in$ \{grouped, depthwise inverted residual bottleneck\}; and channel-scaling $\in$ \{$0.5\times$, $1.0\times$\}. The search-space can be expanded or arbitrarily constrained to known efficient architectures for a device. Each of these $5\times384=1920$ alternative blocks is trained using BKD to complete the Block Library. Once the Block Library is trained, we use the BKD-based ranking metric from DNA\cite{blockwise_nas} to sample a set of architectures uniformly spread over the ranking space. For the \methodname~search space, we finally finetune the sampled networks for 50 epochs starting from the BKD initialization, building an Architecture Library with accuracy targets used to fit the linear accuracy predictor. Typically, 20-30 target networks need to be finetuned to yield good results, see Appendix~\ref{subsec:supplementary/accuracy_predictors}.


In total, including the training of a reference model ($450$ epochs), $450+1920+30\times50=3870$ epochs of training are required to build the accuracy predictor. This is less than $10\times$ the cost of training a single network from scratch to model the accuracy of more than 8 trillion architectures. Subsequently, any architecture can be selected and trained to full accuracy in 50 epochs, starting from the BKD initialization. Similarly, as further discussed in Appendix~\ref{subsec:supplementary/accuracy_predictors}, an accuracy model for MobileNetV3 (1.2$\times$) and EfficientNet-Compressed costs $450+135+20\times50=1585$ epochs, roughly the same as training 4 models from scratch. Although this is a higher cost than OFA \cite{ofa}, it covers a much more diverse search space. OFA requires an equivalent, accounting for dynamic batch sizes~\cite{ofa_repo}, of $180+125+2\times150+4\times150=1205$ epochs of progressive shrinking with backpropagation on a large supernet. BKDNAS \cite{blockwise_nas} requires only $450+16\times20=770$ epochs to build its ranking model, but $450$ epochs to train models from scratch. Other methods like MnasNet~\cite{mnasnet} can handle a similar diversity as \methodname, but typically require an order of magnitude longer search time ($40000$ epochs) \emph{for every deployment scenario}. \methodname~offers MNasNet-level diversity at a 2 orders of magnitude lower search cost. On top of that, BKD epochs are significantly faster than epochs on a full network, as BKD requires only partial computation of the reference model and backpropagation on a single block $B_{n,m}$. Moreover, and in contrast to OFA, all blocks $B_{n,m}$ can be trained in parallel since they are completely independent of each other. Table~\ref{tab:comparison} quantifies the differences in search-time between these approaches.


With the accuracy predictor in place, Pareto-optimal \methodname~models are found for several targets. Figure~\ref{fig:trendlines_overview} shows \methodname~finds networks that outperform the state of the art in terms of the number of parameters, on a simulator targeting tensor compute units in a mobile SoC, on a NVIDIA V100 GPU and on the Samsung S20 GPU. Every predicted Pareto-optimal front is generated using an evolutionary search with NSGA-II \cite{nsga_ii, pymoo} on a population of 100 architectures until convergence. Where applicable, full-architecture hardware measurements are used in the evolutionary loop. Details on measurements and baseline accuracy are given in Appendix~\ref{subsec:supplementary/baselines}. 

\begin{figure}[t]
\centering
\includegraphics[width=.99\linewidth]{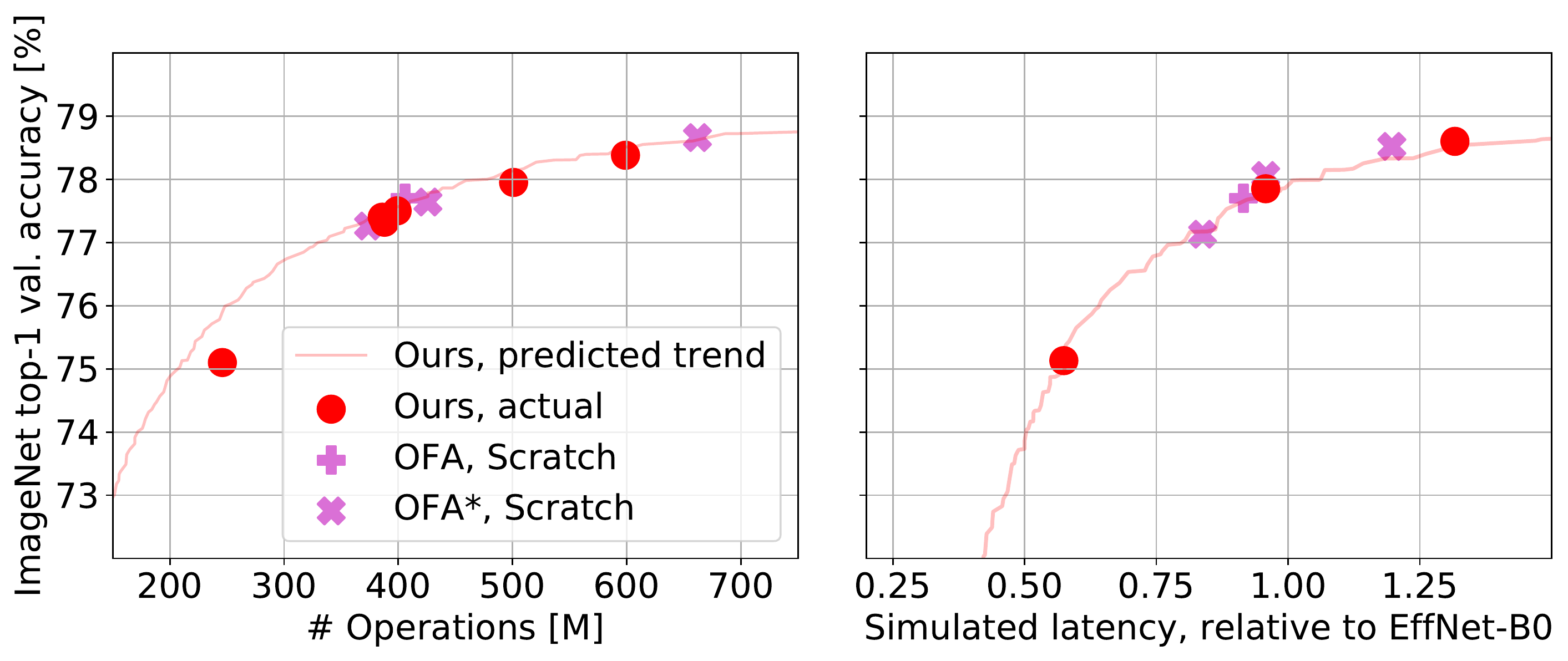}
\caption{DONNA-NAS finds models that are on-par with models found by OFA~\cite{ofa} in the MobileNetV3 (1.2$\times$) search-space. Models are identically trained for fair comparison. OFA* models are found by us using~\cite{ofa_repo} and trained from scratch.}
\label{fig:donna-finds-ofa-models}
\end{figure}

Similarly, Tab.~\ref{tab:mobilenetv3_comparison} and Fig.~\ref{fig:donna-finds-ofa-models} show that \methodname~finds models that are on-par with architectures found by other state-of-the-art methods such as MnasNet~\cite{mnasnet} and OFA~\cite{ofa} in the same spaces. Tab.~\ref{tab:mobilenetv3_comparison} shows DONNA finds models in the MobileNetV3 (1.0$\times$) space that are on par with MobileNetV3~\cite{mbv3} in terms of number of operations, although~\cite{mbv3} is found using expensive MnasNet~\cite{mnasnet}. Fig.~\ref{fig:donna-finds-ofa-models} shows the same for networks found through DONNA in the MobileNetV3 (1.2$\times$) search space, by comparing them to models found through OFA~\cite{ofa} optimized for the same complexity metric and trained with the same hyperparameters. More results for other search spaces are shown in Figure~\ref{fig:supplementary/search-space-overview} in Appendix~\ref{subsec:supplementary/various-search-spaces}. We also visualize Pareto-optimal \methodname~models for different platforms in Appendix~\ref{subsec:supplementary/model-visualizations}.

\begin{table}[]
\small
\def\arraystretch{0.9}%
\caption{\label{tab:mobilenetv3_comparison} \methodname~finds similar models to MobileNetV3~\cite{mbv3} in the MobileNetV3 (1.0$\times$) space.}
\centering
\begin{tabular}{l|c|c}
\multirow{2}{*}{Network} &  Number of & ImageNet   \\
& Operations [M]& val top-1 [\%] \\
\hline
MobileNetV3~\cite{mbv3} & 232 & 75.77@600\cite{rw_imagenet} \\ 
Ours (MobNetV3 $1.0\times$)& 242 & 75.75@50 \\ 
\end{tabular}
\end{table}

\subsubsection{Search-Space Extension and Exploration}
\label{subsubsec:experiments/search-space-design}
The \methodname~approach can also be used for \textit{rapid search space extension and exploration}. Using \methodname, a designer can quickly determine whether the search space should be extended or constrained for optimal performance.

\begin{figure}[t]
\centering
\includegraphics[width=1.00\linewidth]{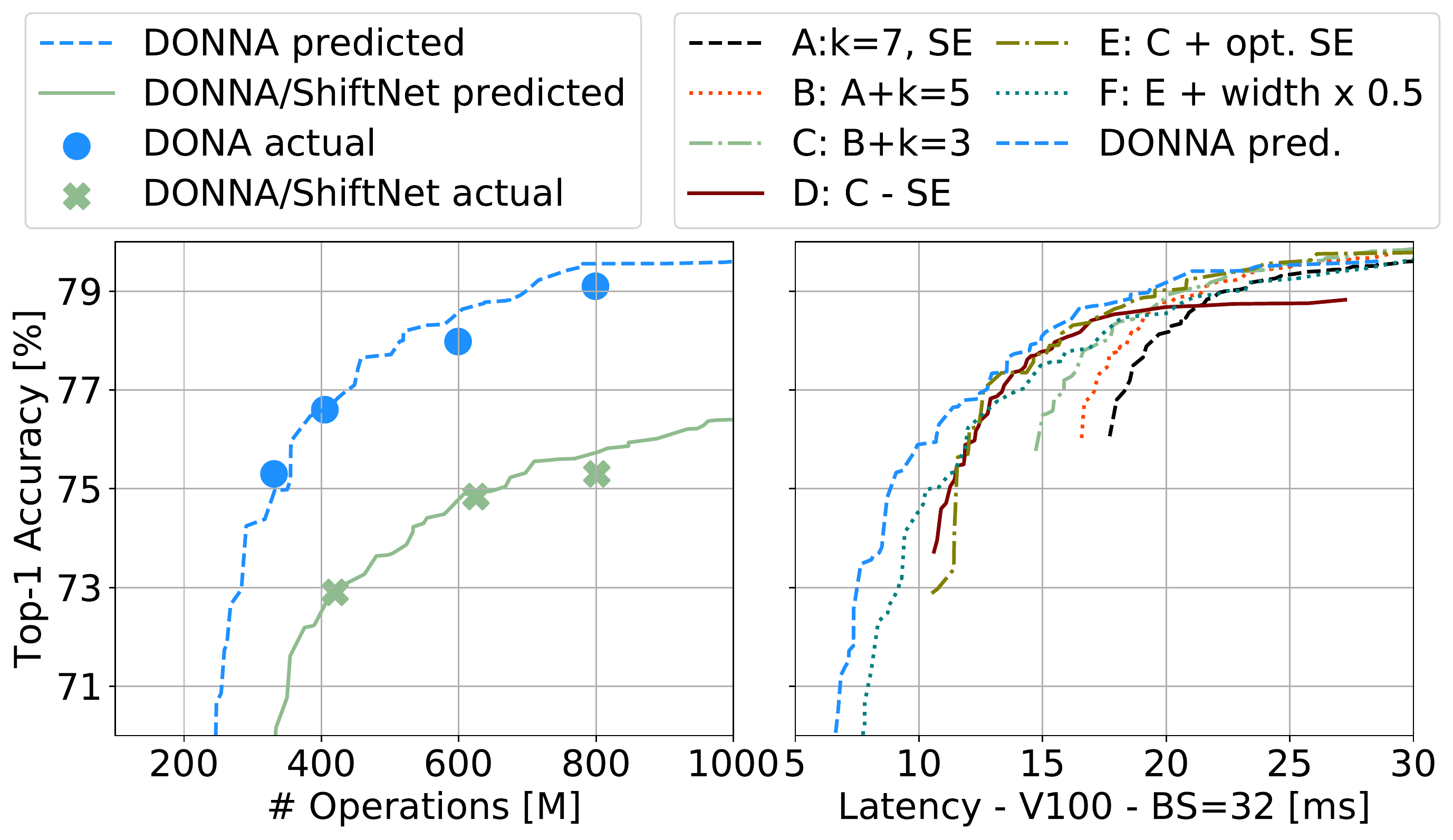}
\caption{(left) An accuracy predictor for \methodname~generalizes to an unseen space with ShiftNets \cite{shiftnet}, without using ShiftNets to train the predictor. (right) Rapid, model-driven exploration of models within the original \methodname~search-space on a V100 GPU. The figure illustrates the necessity of a diverse search space, achieving up to 25$\%$ latency gains when attention can be chosen optimally (line E vs C).}
\label{fig:search_space_analysis}
\end{figure}

Such \textit{extension} is possible because the \methodname~accuracy predictor generalizes to previously unseen architectures, without having to extend the Architecture Library. This is illustrated in Fig.~\ref{fig:linear_predictor}(left), showing the DONNA predictor achieves good quality, in line with the original test set, on a ShiftNet-based test set of architectures.
Figure~\ref{fig:search_space_analysis}(left) further illustrates this extrapolation works by showing the confirmed results of a search for the ShiftNet space. Note how the trendline predicts the performance of full Pareto optimal ShiftNets even though the predictor is created without any ShiftNet data. Here, ShiftNets are our implementation, with learned shifts per group of 32 channels as depthwise-separable replacement. These generalization capabilities are obtained because the predictor only uses quality metrics as an input without requiring any structural information about the replacement block. 
This feature is a major advantage of~\methodname~compared to OFA~\cite{ofa} and other methods where the predictor cannot automatically generalize to completely different layer-types, or to blocks of the same layer-type with parameters (expansion rate, kernel size, depth, ...) outside of the original search space. Appendix~\ref{sec:supplementary/qnas} illustrates such extension can also be used to model accuracy of lower precision quantized networks.



This prototyping capability is also showcased for the \methodname~search space on a V100 GPU in Figure~\ref{fig:search_space_analysis}(right). Here we interpolate, using the original accuracy predictor for \textit{exploration}. In doing this, Fig.~\ref{fig:search_space_analysis} shows search-space diversity is crucial to achieve good performance. Especially the impact of optimally adding SE-attention~\cite{squeeze_excite} is very large, predicting a 25$\%$ speedup at 76$\%$ accuracy (line C vs D), or a 1$\%$ accuracy boost at 26ms (line E vs D). 
Every plotted line in Figure~\ref{fig:search_space_analysis} (right) is a predicted Pareto-optimal. A baseline (A) considers SE/Swish in every block and k $\in$ \{7\}, expand $\in$ \{3,4,6\} and depth $\in$ \{2,3,4\}. Other lines show results for search spaces built starting from (A), e.g. (B) considers k $\in$ \{5,7\}, (C) k $\in$ \{3,5,7\}, (D) removes SE/Swish, (E) allows choosing optimal placement of SE/Swish, (F) adds a channel-width multiplier.

\subsubsection{Model Compression}
\label{subsubsec:experiments/compression}

\begin{figure}[b]
\centering
\includegraphics[width=.99\linewidth]{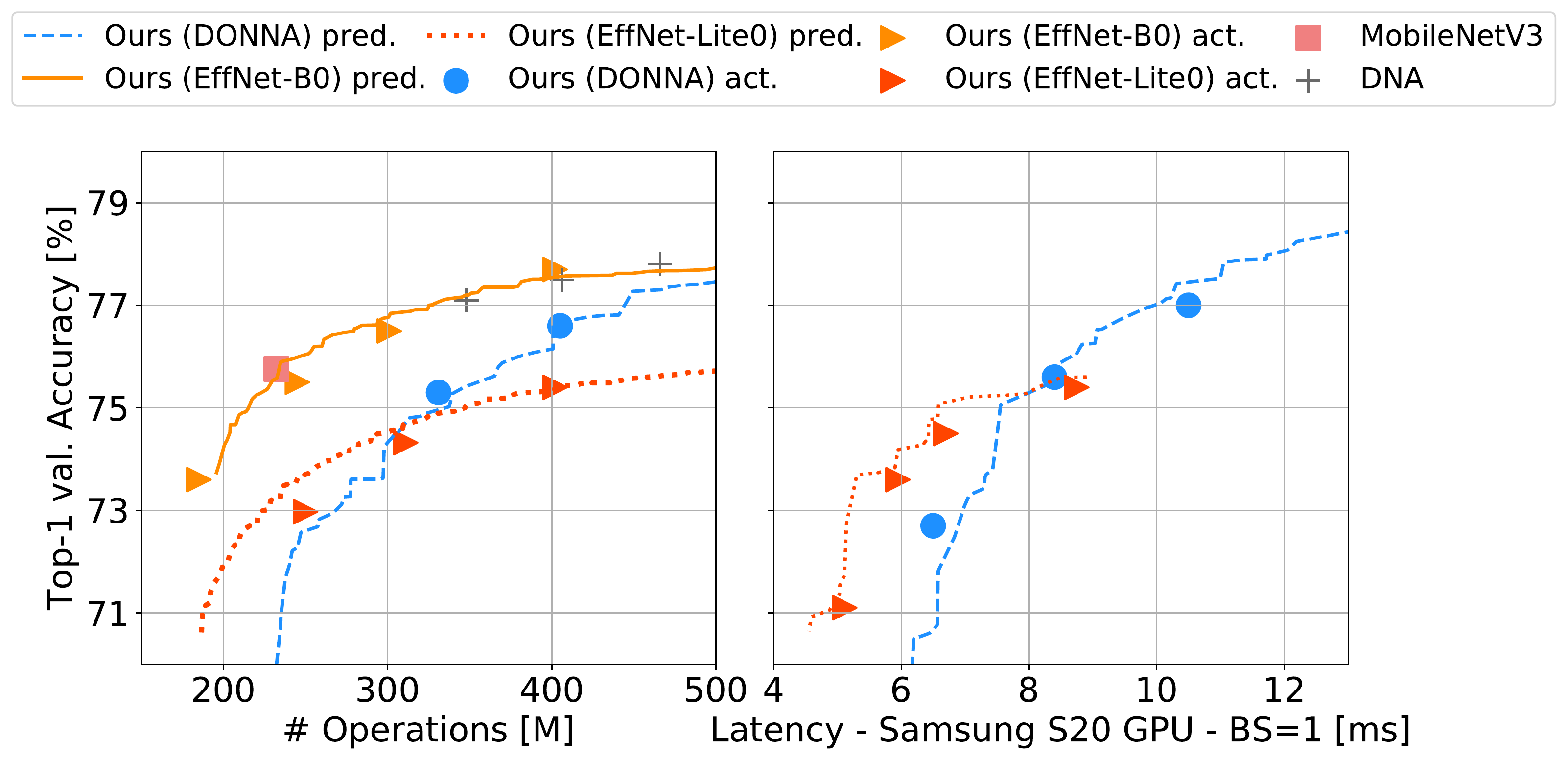}
\caption{Compressing EfficientNet-B0 for two targets.}
\label{fig:efficientnet}
\end{figure}

\methodname~is also used for \textit{hardware-aware compression of existing neural architectures} into faster, more efficient versions. \methodname~can do compression not just in terms of the number of operations, as is common in literature, but also for different devices.
This is useful for a designer who has prototyped a network for their application and wants to run it efficiently on many different devices with various hardware and software constraints.  Figure~\ref{fig:efficientnet} shows how EfficientNet-B0 can be compressed into networks that are $10\%$ faster than MnasNet~\cite{mnasnet} on the Samsung S20 GPU.

In the \methodname~compression pipeline, the EfficientNet search space splits EfficientNet-B0 into 5 blocks and uses it as the reference model. Every replacement block $B_{n,m}$ considered in compression is smaller than the corresponding reference block. $1135$ epochs of training are spent in total to build an accuracy predictor: 135 blocks are trained using BKD, and 20 architectures are trained for 50 epochs as prediction targets, a cost equivalent to the resources needed for training 3 networks from scratch. 
Figure~\ref{fig:efficientnet} shows \methodname~finds a set of smaller, Pareto optimal versions of EfficientNet-B0 both in the number of operations and on-device. These are on-par with MobileNetV3~\cite{mbv3} in the number of operations and $10\%$ faster than MnasNet~\cite{mnasnet} on device. For Samsung S20, the accuracy predictor is calibrated, as these models have no SE and Swish in the head and stem as in the EfficientNet-B0 reference.

Similarly, DONNA can be used to optimally compress Vision Transformers (ViT~\cite{dosovitskiy2020image}), see Appendix~\ref{sec:supplementary/vit}.

\subsection{Object Detection on MS-COCO}
\label{subsec:experiments/detection}
\begin{figure}[t]
\centering
\includegraphics[width=.99\linewidth]{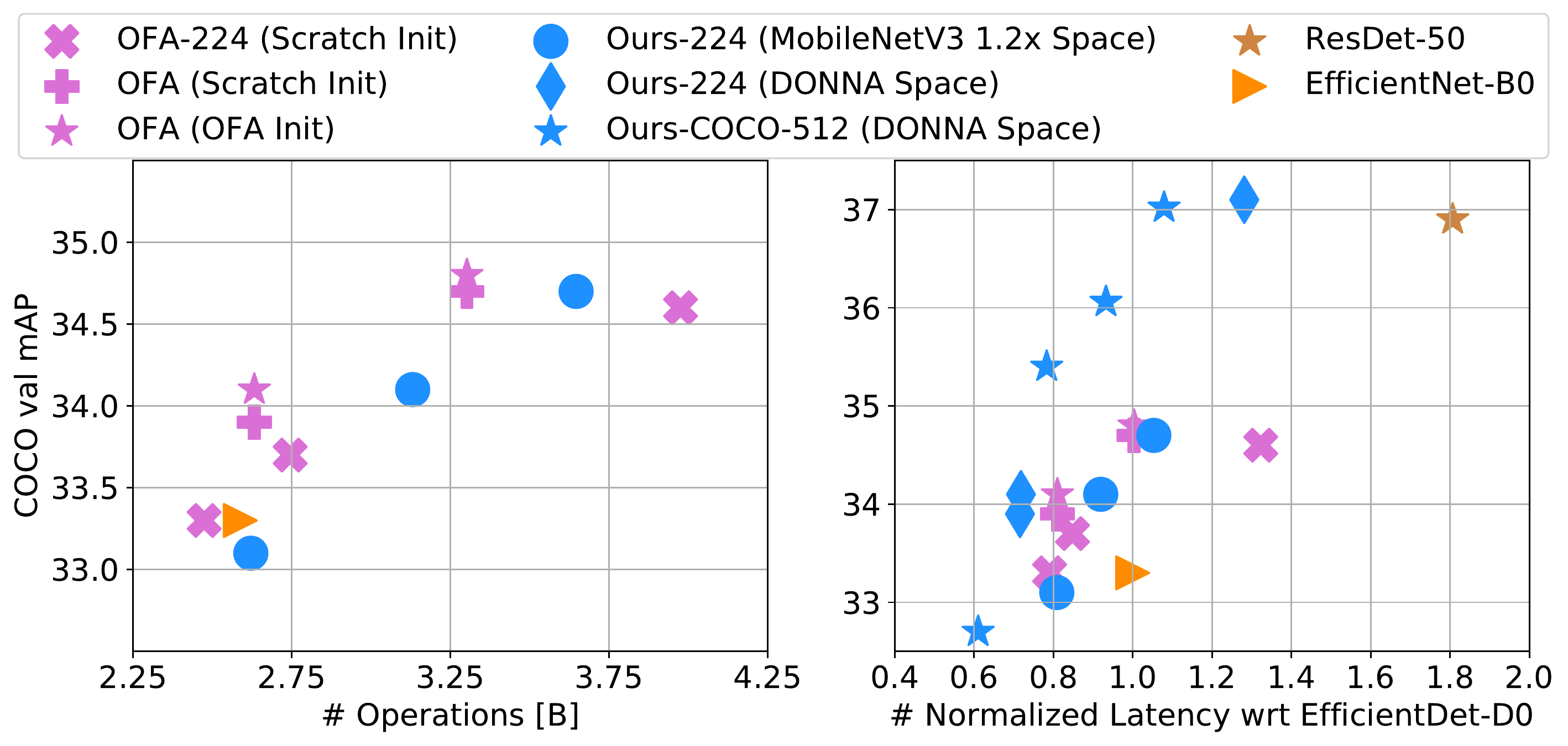}
\caption{Object detection performance of \methodname~backbones, either searched on ImageNet and transferred to COCO (Ours-224), or searched directly on MS COCO (Ours-COCO-512). In the \methodname~search space, our solution has up to $2.4\%$ higher mAP at the same latency as the OFA models.}
\label{fig:detection}
\end{figure}
The \methodname~architectures transfer to other tasks such as object detection on MS COCO \cite{coco}.
To this end, we use the EfficientDet-D0 \cite{efficientdet} detection architecture, replacing its backbone with networks optimized through the DONNA pipeline. For training, we use the hyperparameters given in \cite{rw_coco}.
The EfficientDet-D0 initialization comes from \cite{rw_imagenet}.


Figure~\ref{fig:detection} shows the results of multiple of such searches. First, we optimize backbones on ImageNet in the MobileNetV3 (1.2$\times$) and ~\methodname~spaces (ours-224), targetting both the number of operations (left) and latency on a simulator targeting tensor compute units. In this case, the input resolution is fixed to $224\times224$. The backbones are first finetuned on ImageNet and then transferred to MS-COCO. Second, we apply the DONNA pipeline directly on the full DONNA-det0 architecture, building an accuracy predictor for MS-COCO. We optimize only the backbone and keep the BiFPN head fixed (Ours-COCO-512). In this case, the resulting networks are directly finetuned on MS-COCO, following the standard DONNA-flow.
For OFA \cite{ofa}, we consider two sets of models. The first set consists of models optimized for the number of operations (FLOP) with varying input resolution coming directly from the OFA repository \cite{ofa_repo}. The second set of models, which we identify by `OFA-224', are obtained by us with the same tools \cite{ofa_repo}, but with the input resolution fixed to $224\times224$. This makes the OFA-224 search space the same as our MobileNetV3 (1.2$\times$) up to the layerwise-vs-blockwise distinction. In the first experiment, we initialize the OFA backbone with weights from progressive shrinking released in \cite{ofa_repo}. In the second experiment, we initialize the OFA backbone with from-scratch trained weights on ImageNet using hyperparameters from \cite{rw_imagenet}. After such initialization, the networks are transferred to object detection for comparison. The comparison of the two experiments shows the benefit of OFA-style training is limited after transfer to a downstream task (See Fig.~\ref{fig:detection}.) The gap between OFA-style training and training from scratch, which is up to $1.4\%$ top-1 on ImageNet, decreases to $0.2\%$ mAP on COCO, reducing its importance. We discuss this point further in Appendix~\ref{sec:supplementary/transfer}.

In comparing with \methodname~models, we make three key observations. First, models transferred after a search using \methodname~are on-par or better than OFA-224 models for both operations and latency. Second, models transferred from the \methodname~space outperform OFA models up to $2.4\%$ mAP on the validation set in latency. Third, best results are achieved when applying DONNA directly to MS COCO. 
\section{Conclusion} 
\label{sec:conclusion}
In this work, we present \methodname, a novel approach for rapid scenario-aware NAS in diverse search spaces. 
Through the use of a model accuracy predictor, built through knowledge distillation, \methodname~finds     
state-of-the-art networks for a variety of deployment scenarios: in terms of number of parameters and operations, and in terms of latency on Samsung S20 and the Nvidia V100 GPU. In ImageNet classification, architectures found by \methodname~are 20\% faster than EfficientNet-B0 and MobileNetV2 on V100 at similar accuracy and 10\% faster with 0.5\% higher accuracy than MobileNetV2-1.4x on a Samsung S20 smartphone. In object detection, \methodname~finds networks with up to $2.4\%$ higher mAP at the same latency compared to OFA. Furthermore,  this pipeline can be used for quick search space extensions (e.g. adding ShiftNets) and exploration, as well as for on-device network compression.


{\small
\bibliographystyle{ieee_fullname}
\bibliography{references}
}

\newpage

\appendix
\twocolumn[{%
 \centering
 \LARGE \textbf{Appendix}\\[1.5em]
}]

\section{Experimental Details}
\label{sec:supplementary}

\subsection{Hyperparameters for training and distillation}
\label{subsec:supplementary/training_hyperparameters}
All reference models for each search space are \textbf{trained from scratch} for 450 epochs on 8 GPUs up to state-of-the-art accuracy using the hyperparameters given in \cite{rw_imagenet} for EfficientNet-B0 \cite{efficientnet}. More specifically, we use a total batch size of 1536 with an initial learning rate of 0.096, RMSprop with momentum of 0.9, RandAugment data augmentation \cite{randaugment}, exponential weight-averaging, dropout \cite{dropout} and stochastic depth \cite{stochastic_depth} of 0.2, together with a learning rate decay of 0.97 every 2.4 epochs. 

\textbf{Blockwise knowledge distillation (BKD)} is done by training every block for a single epoch. During this epoch, we apply a cosine learning rate schedule \cite{cosine} considering 20 steps, an initial learning rate of 0.01, a batch size of 256, the Adam \cite{kingma2014adam} optimizer, and random cropping and flipping as data augmentation.

\textbf{Finetuning} is done via end-to-end knowledge distillation (EKD) by using hard ground truth labels and the soft labels of the reference model, see Figure~\ref{fig:bkd_overview}(b). We use the same hyperparameters used for training from scratch with the following changes: a decay of 0.9 every 2 epochs, the initial learning rate divided by 5 and no dropout, stochastic depth nor RandAugment. Depending on the reference model and the complexity of the search space, finetuning achieves full from-scratch accuracy in 15-50 epochs, see Figure~\ref{fig:supplementary/finetuning-speed}.

\subsection{Hardware measurements}
\label{subsec:supplementary/hardware_measurements}

All complexity measurements used throughout the text, either hardware-aware or hardware-agnostic, are gathered as follows:
\begin{itemize}
    \item \textbf{Nvidia V100 GPU} latency measurements are done in Pytorch 1.4 with CUDNN 10.0. In a single loop, 20 batches are sent to GPU and executed, while the GPU is synced before and after every iteration. The first 10 batches are treated as a warm-up and ignored; the last 10 are used for measurements. We report the fastest measurement as the latency.
    \item Measurements on the \textbf{Samsung S20 GPU} are always done with a batch-size of 1, in a loop running 30 inferences, after which the system cools down for 1 minute. The average latency is reported.
    \item The \textbf{number of operations} and \textbf{number of parameters} are measured using the ptflops framework (https://pypi.org/project/ptflops/).
    \item Latency measurement on the \textbf{simulator} targeting tensor compute units is done with a batch-size of 1. We report the fastest measurement as latency.
\end{itemize}

All complexity metrics for the reference models shown throughout the text are measured using this same setup.

\subsection{Accuracy of baseline models}
\label{subsec:supplementary/baselines}
Accuracy is taken to be the highest reported in~\cite{rw_imagenet}, the highest reported in the paper, or trained from scratch using the EfficientNet-B0 hyperparameters used in the \cite{rw_imagenet} repository, see Table~\ref{tab:supplementary/baseline_models}. This is the case for EfficientNet-B0 (our training), MobileNetV2, MnasNet, SPNASNet and FBNet. OFA/Scratch is the  
``flops@389M\_top1@79.1\_finetune@75'' model from ~\cite{ofa_repo} trained from scratch using the hyperparameters used for EfficientNet-B0 in \cite{rw_imagenet}.  Note that these baselines are competitive. MobileNetV2 for example, typically has an accuracy of around 72$\%$, while the training in~\cite{rw_imagenet} pushes that to 73$\%$. ResNet50 is typically at 76$\%$, but reaches 79$\%$ using the training proposed in \cite{rw_imagenet}. ProxylessNas~\cite{proxylessnas} and DNA's~\cite{blockwise_nas} accuracy is taken from their respective papers.

\begin{table}[]
\small
\caption{\label{tab:supplementary/baseline_models} Top-1 ImageNet validation accuracy of architectures used throughout the text, with the references indicating the source for the listed accuracy. * are models found by us using OFA~\cite{ofa_repo} for a specific target complexity metric.}
\centering
\begin{tabular}{l|c|c}
Architecture & ImageNet Top-1 [$\%$] & Reference \\ 
\hline
EfficientNet-B0 & 77.7 & Ours, using \cite{rw_imagenet} \\
SPNASNet-100 & 74.084 & From \cite{rw_imagenet} \\
MNasNet-B1-1.0$\times$ & 74.658 & From \cite{rw_imagenet} \\
MNasNet-A1-1.0$\times$ & 75.448 & From \cite{rw_imagenet} \\
MNasNet-A1-1.4$\times$ & 77.2 & From \cite{mnasnet} \\
FBNet-C-100 & 78.124 & From \cite{rw_imagenet} \\
MobileNetV2 (1.0x) & 72.970 & From \cite{rw_imagenet} \\
MobileNetV2 (1.4x) & 76.516 & From \cite{rw_imagenet} \\
MobileNetV3 (Large) & 75.766 & From \cite{rw_imagenet} \\
ProxyLessNas CPU & 75.3 & From \cite{proxylessnas} \\
ProxyLessNas GPU & 75.1 & From \cite{proxylessnas}\\
ProxyLessNas Mobile & 74.6 & From \cite{proxylessnas} \\
ResNet34 & 75.1 & From \cite{rw_imagenet} \\
ResNet50 & 79.0 & From \cite{rw_imagenet} \\
OFA/Scratch & 77.5 & Ours, with \cite{rw_imagenet}\\
OFA-flops-A* & 77.3 & Ours, with \cite{rw_imagenet}\\
OFA-flops-B* & 77.5 & Ours, with \cite{rw_imagenet}\\
OFA-flops-C* & 78.6 & Ours, with \cite{rw_imagenet}\\
OFA-sim-A* & 77.1 & Ours, with \cite{rw_imagenet}\\
OFA-sim-B* & 78.1 & Ours, with \cite{rw_imagenet}\\
OFA-sim-C* & 78.5 & Ours, with \cite{rw_imagenet}\\
DNA-A & 77.1 & From \cite{blockwise_nas} \\
DNA-B & 77.5 & From \cite{blockwise_nas} \\
DNA-C & 77.8 & From \cite{blockwise_nas} \\
DNA-D & 78.4 & From \cite{blockwise_nas} \\

\end{tabular}
\end{table}


\subsection{Comments on Accuracy Predictors}
\label{subsec:supplementary/accuracy_predictors}

\subsubsection{Size of the Architecture Library}
\label{subsubsec:supplementary/architecture_library}
Tables~\ref{tab:supplementary/library-size-mobnetv3} and~\ref{tab:supplementary/library-size-donna} show the impact of the size of the Architecture Library used to fit the linear predictor. The tables show how performance varies on a test set of finetuned models for the MobileNetV3 (1.2$\times$) and \methodname~search spaces, respectively. Note how the ranking quality, as measured by Kendall-Tau (KT)~\cite{kendall1938new}, is always better in this work than in DNA~\cite{blockwise_nas}. On top of that, DNA~\cite{blockwise_nas} only ranks models within the search space and does not predict accuracy itself. Another metric to estimate the accuracy predictor's quality is the Mean-Squared-Error (MSE) in terms of predicted top-1 accuracy on the ImageNet validation set. Note that for the MobileNetV3 (1.2$\times$) search space, 20 target accuracies are sufficient for a good predictor, as shown in Table~\ref{tab:supplementary/library-size-mobnetv3}. We use the same amount of targets for the EfficientNet-B0, MobilenetV3 (1.0$\times$) and ProxylessNas (1.3$\times$) search spaces. For the \methodname~search space, we use 30 target accuracies, see Table~\ref{tab:supplementary/library-size-donna}. Note that the linear accuracy predictor can improve overtime, whenever the Architecture Library is expanded. As predicted Pareto-optimal architectures are finetuned to full accuracy, those results can be added to the library and the predictor can be fitted again using this extra data.

\begin{table}[]
\small
\caption{\label{tab:supplementary/library-size-mobnetv3} Ranking quality for MobileNetV3 (1.2$\times$) using \methodname, as function of the size of the Architecture Library. `X'T indicates that `X' targets were used to fit the predictor.}
\centering
\begin{tabular}{l|c|c|c|c|c}
Metric & DNA \cite{blockwise_nas} & 10T & 20T & 30T & 40T \\ 
\hline
Kendall-Tau \cite{kendall1938new} & 0.74 & 0.79 & 0.79 & 0.8 & 0.82 \\
MSE [top-1\%] & NA & 0.07 & 0.09 & 0.09 & 0.08 \\
\end{tabular}
\end{table}

\begin{table}[]
\small
\caption{\label{tab:supplementary/library-size-donna} Ranking quality for \methodname, as a function of the size of the Architecture Library. `X'T indicates that `X' targets were used to fit the predictor.}
\centering
\begin{tabular}{l|c|c|c|c|c}
Metric & DNA \cite{blockwise_nas} & 10T & 20T & 30T & 40T \\ 
\hline
Kendall-Tau \cite{kendall1938new} & 0.77 & 0.87  & 0.87 & 0.9 & 0.9  \\
MSE [top-1\%] & NA & 0.28  & 0.18 & 0.2 & 0.19  \\
\end{tabular}
\end{table}
\subsubsection{Choice of Quality Metrics}
\label{subsubsec:supplementary/comparing_quality_metrics}
Apart from using the Noise-To-Signal-Power-Ratio (NSR) (See Section~\ref{sec:search_method}), other quality metrics can be extracted and used in an accuracy predictor as well. All quality metrics are extracted on a held-out validation set, sampled from the ImageNet training set, which is different from the default ImageNet validation set in order to prevent overfitting. Three other types of quality metrics are considered on top of the  metric described in equation~\ref{eq:loss}: one other block-level metric based on L1-loss and two network-level metrics. The block-level metric measures the normalized L1-loss between ideal feature map $Y_n$ and the block $B_{n,m}$'s output feature map $\bar{Y}_{n,m}$. It can be described as the Noise-to-Signal-Amplitude ratio:
\begin{equation}
\label{eq:l1_alternative}
\mathcal{L}(W_{n,m};Y_{n-1},Y_n) = \frac{1}{C}\sum_{c=0}^{C} \frac{\|Y_{n,c}-\bar{Y}_{n,m,c}\|_1}{\sigma_{n,c}}
\end{equation}
The two network-level metrics are the loss and top-1 accuracy extracted on the separate validation set. The network-level metrics are derived by replacing only block $B_n$ in the reference model with the block-under-test $B_{n,m}$ and then validating the performance of the resulting network. Table~\ref{tab:supplementary/accuracy_model} compares the performance of the 4 different accuracy predictors built on these different styles of features. Although they are conceptually different, they all lead to a very similar performance on the test set with NSR outperforming the others slightly. Because of this, the NSR metric from equation~\ref{eq:loss} is used throughout the text.
%
\begin{table}[]
\small
\caption{\label{tab:supplementary/accuracy_model} Comparing different quality metrics: NSR (Equation~\ref{eq:loss}), L1, network-level loss and top-1 accuracy for \methodname.}
\centering
\begin{tabular}{l|c|c|c|c|c}
Ranking Metric & DNA \cite{blockwise_nas} & NSR & L1 & Loss & Top-1 \\ 
\hline
Kendall-Tau \cite{kendall1938new} & 0.77 & 0.9  & 0.89 & 0.89 & 0.88  \\
MSE [top-1\%] & NA & 0.19  & 0.23 & 0.41 & 0.44  \\
\end{tabular}
\end{table}

\subsubsection{Accuracy predictors for different search-spaces}
\label{subsubsec:supplementary/accuracy_predictors_different_searchspaces}
Similar to the procedures discussed in section~\ref{sec:search_method}, accuracy models are built for different reference architectures in different search spaces: EfficientNet-B0, MobileNetV3 (1.0$\times$), MobileNetV3 (1.2$\times$) and ProxyLessNas (1.3$\times$). The performance of these models is illustrated in Table~\ref{tab:supplementary/other_predictors}. Note that we can generate reliable accuracy predictors for all of these search spaces, with very high Kendall-Tau ranking metrics and low MSE on the prediction. The Kendall-Tau value on the MobileNetV3 ($1.2\times$) search space is lower than the others, as the test set is larger for this space than for the others. The model is still reliable, as is made apparent by the very low MSE metric.
\begin{table}[]
\small
\caption{\label{tab:supplementary/other_predictors} Comparing the quality of accuracy predictors for different search spaces. Predicted accuracy is the top-1 validation accuracy on ImageNet.}
\centering
\begin{tabular}{l|c|c}
Search-Space &  Kendall Tau\cite{kendall1938new}  & MSE [top-1\%] \\ 
\hline
\methodname & 0.9 & 0.19\\ 
EfficientNet-B0 & 0.91 & 0.15\\ 
MobileNetV3 (1.0$\times$) &  0.97 & 0.13 \\ 
MobileNetV3 (1.2$\times$) &  0.82 & 0.08 \\ 
ProxyLessNas (1.3$\times$) &  0.95 & 0.04 \\ 
\end{tabular}
\end{table}

\subsubsection{Ablation on accuracy predictor}
\label{subsubsec:supplementary/accuracy_predictor_ablation}
Throughout this work, we use the Ridge regression from scikit-learn~\cite{scikit-learn} as an accuracy predictor. Other choices can also be valid, although the Ridge regression model has proven stable across our experiments. Table~\ref{tab:supplementary/predictor_ablation_pt2} compares a non-exhaustive list of accuracy predictors from scikit-learn and their performancde on the DONNA architectural test set.
\begin{table}[]
\small
\caption{\label{tab:supplementary/predictor_ablation_pt2} Comparing the quality of different accuracy predictors for the DONNA search space. }
\centering
\begin{tabular}{l|c|c}
Predictor &  Kendall Tau\cite{kendall1938new}  & MSE [top-1\%] \\ 
\hline
\methodname & 0.9 & 0.19\\ 
Ridge & 0.91 & 0.15\\ 
Adaboost &  0.89 & 0.16 \\ 
Lasso &  0.91 & 0.20 \\ 
SVM &  0.86 & 1.84 \\ 
Grad. Boosting Ensemble & 0.86 & 0.53 \\
LARS & 0.16 & 15.9 \\
BaggingRegressor & 0.89 & 0.23 \\
\end{tabular}
\end{table}

\subsection{Finetuning speed}
\label{subsec:supplementary/finetuning-speed}
\begin{figure}[t]
\centering
 \includegraphics[width=.99\linewidth]{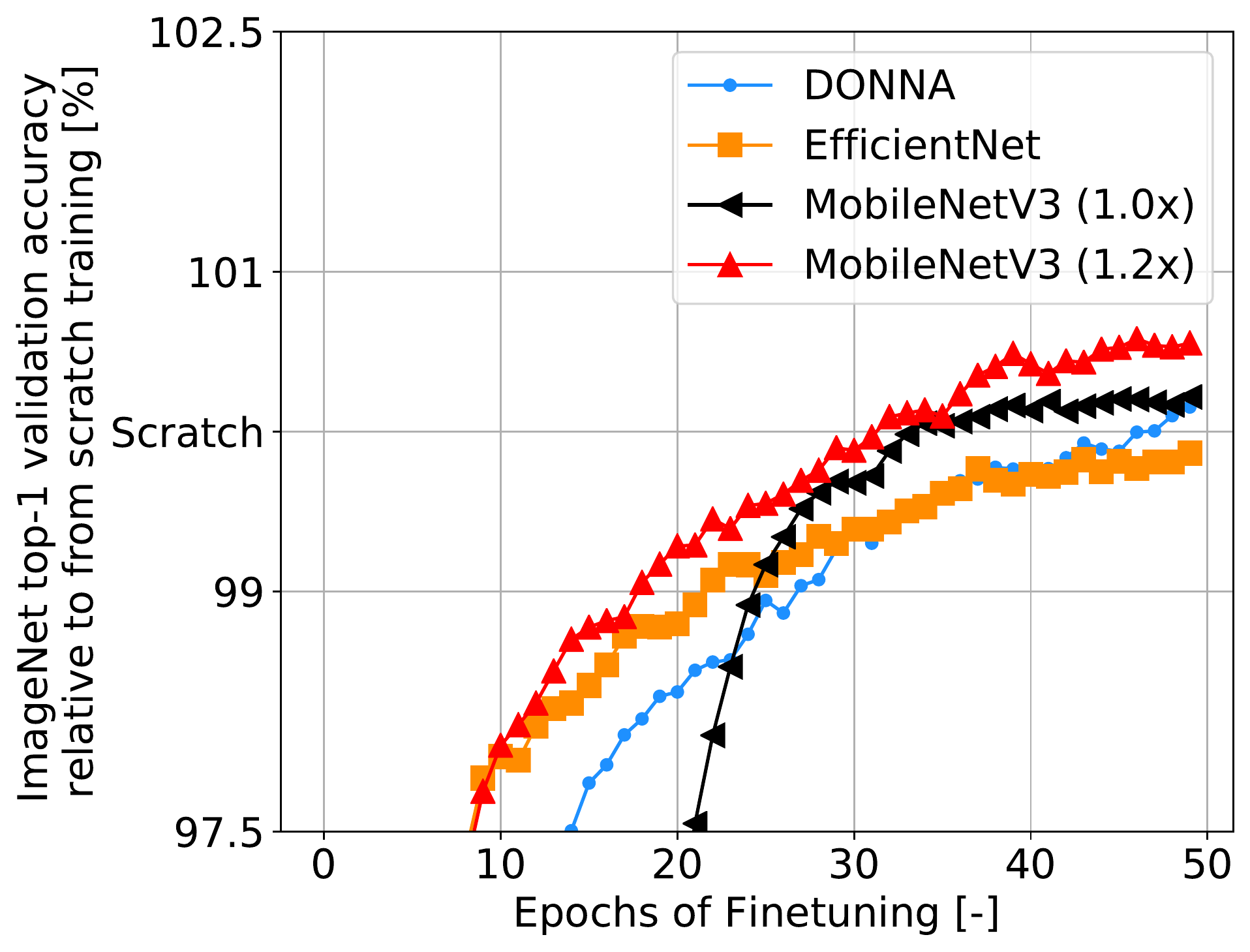}
\caption{Speed at which BKD-initialized subsampled models can be finetuned for different search spaces. Models in ~\methodname, EfficientNet and converge to the accuracy of 450-epoch from scratch training in less than 50 epochs using the BKD initialization point, a $9\times speedup$.}
\label{fig:supplementary/finetuning-speed}
\end{figure}

Depending on the search space's complexity, the used reference model in BKD, and the teacher in end-to-end knowledge distillation (EKD), finetuning can be faster or slower in terms of epochs. We always calibrate the finetuning process to be on-par with training from scratch for a fair comparison, but networks can be trained longer for even better results. With the hyperparameters for EKD given in Appendix~\ref{subsec:supplementary/training_hyperparameters}, Figure~\ref{fig:supplementary/finetuning-speed} shows that finetuning rapidly converges to from-scratch training accuracy for a set of sub-sampled models in different search spaces. Typically, 50 epochs are sufficient for most of the examples. Finetuning speed also depends on the final accuracy of the sub-sampled model. With an accuracy very close to the accuracy of the reference model, larger models typically converge slower using EKD than smaller models with a lower accuracy. For the smaller models, the teacher's guidance dominates more, which leads to faster finetuning.

\begin{figure*}[t]
\centering
 \includegraphics[width=.99\linewidth]{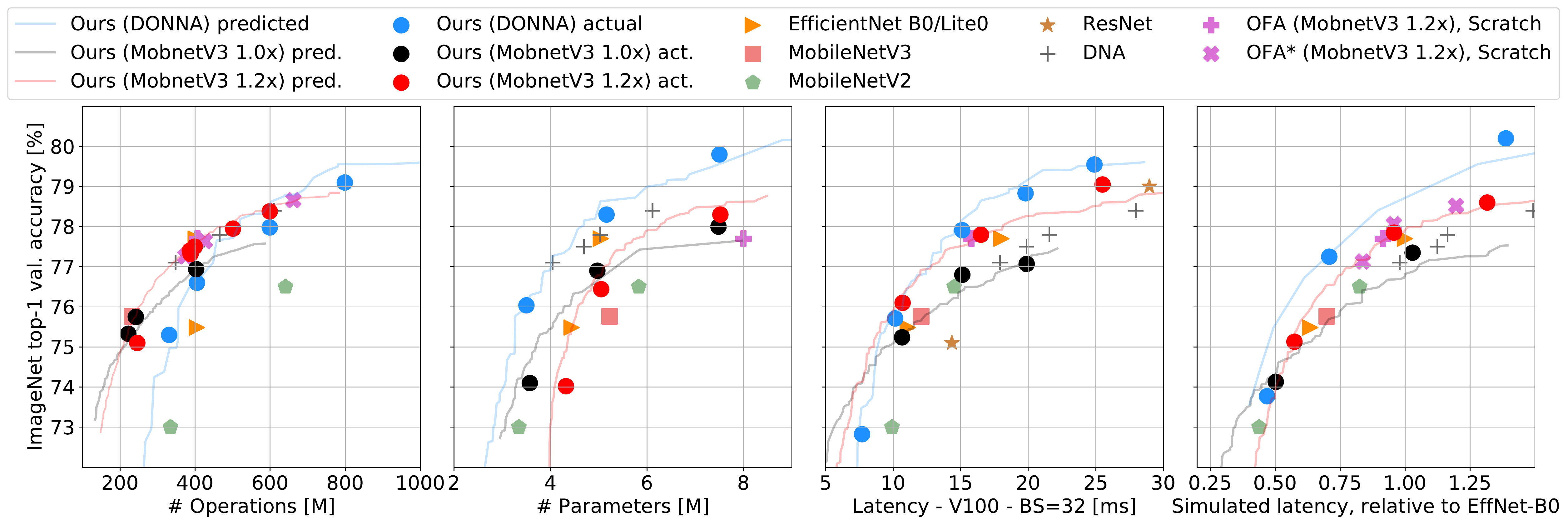}
\caption{Trendlines and models found by \methodname~optimizing for the number of operations (left), the number of parameters (mid left), inference time on an Nvidia V100 GPU (mid right) and a simulator targetting tensor compute units (right). Best viewed in color. This Figure shows the DONNA pipeline finds models of the same quality as OFA~\cite{ofa} when searching in the same search space and optimizing for the same complexity metric (left, right). Second, it shows networks in the DONNA search-space outperform models in the MobileNetV3-$1.0\times$ and MobileNetV3-$1.2\times$ spaces when targeting the number of parameters, or latency on the discussed hardware platforms. When optimizing for the number of operations, the MobileNetV3-style spaces outperform the DONNA space at accuracies lower than $79\%$.}
\label{fig:supplementary/search-space-overview}
\end{figure*}
\subsection{Models for various search-spaces}
\label{subsec:supplementary/various-search-spaces}
Figure~\ref{fig:supplementary/search-space-overview} illustrates predicted and measured performance of \methodname~models in terms of number of operations, number of parameters, on an Nvidia V100 GPU and on a simulator targeting tensor operations in a mobile SoC. On top of this, predicted Pareto curves for a variety of other search-spaces are shown: MobileNetV3 (1.0$\times$) and MobileNetV3 (1.2$\times$). For these other search-spaces, we perform predictor-based searches in each of the scenarios, illustrating their respective predicted Pareto-optimal trendlines. The quality of these predictors is given in Table~\ref{tab:supplementary/other_predictors}. For the extra search spaces, some optimal models have been finetuned to verify the predicted curve's validity. For every search space, the same accuracy predictor is used across all scenarios. 

MobileNetV3 (1.0$\times$) and MobileNetV3 (1.2$\times$) are confirmed in terms of number of operations in Figure~\ref{fig:supplementary/search-space-overview} (mid-left). ProxyLessNass (1.3$\times$) is confirmed on an Nvidia V100 GPU in Figure~\ref{fig:supplementary/search-space-overview} (mid-right). In the MobileNetV3 ($1.0\times$) space, we find networks that are on-par with the performance of MobileNetV3~\cite{mbv3} in terms of accuracy for the same number of operations, which validates that \methodname~can find the same optimized networks as other methods in the same or similar search spaces. Note that the \methodname~outperforms all other search spaces on hardware platforms and in terms of number of parameters, which motivates our choice to introduce the new design space. The \methodname~space is only outperformed in terms of Pareto-optimality when optimizing for the number of operations, a proxy metric.

\section{Model Transfer Study} 
\label{sec:supplementary/transfer}
\begin{figure}[t]
\centering
\includegraphics[width=.99\linewidth]{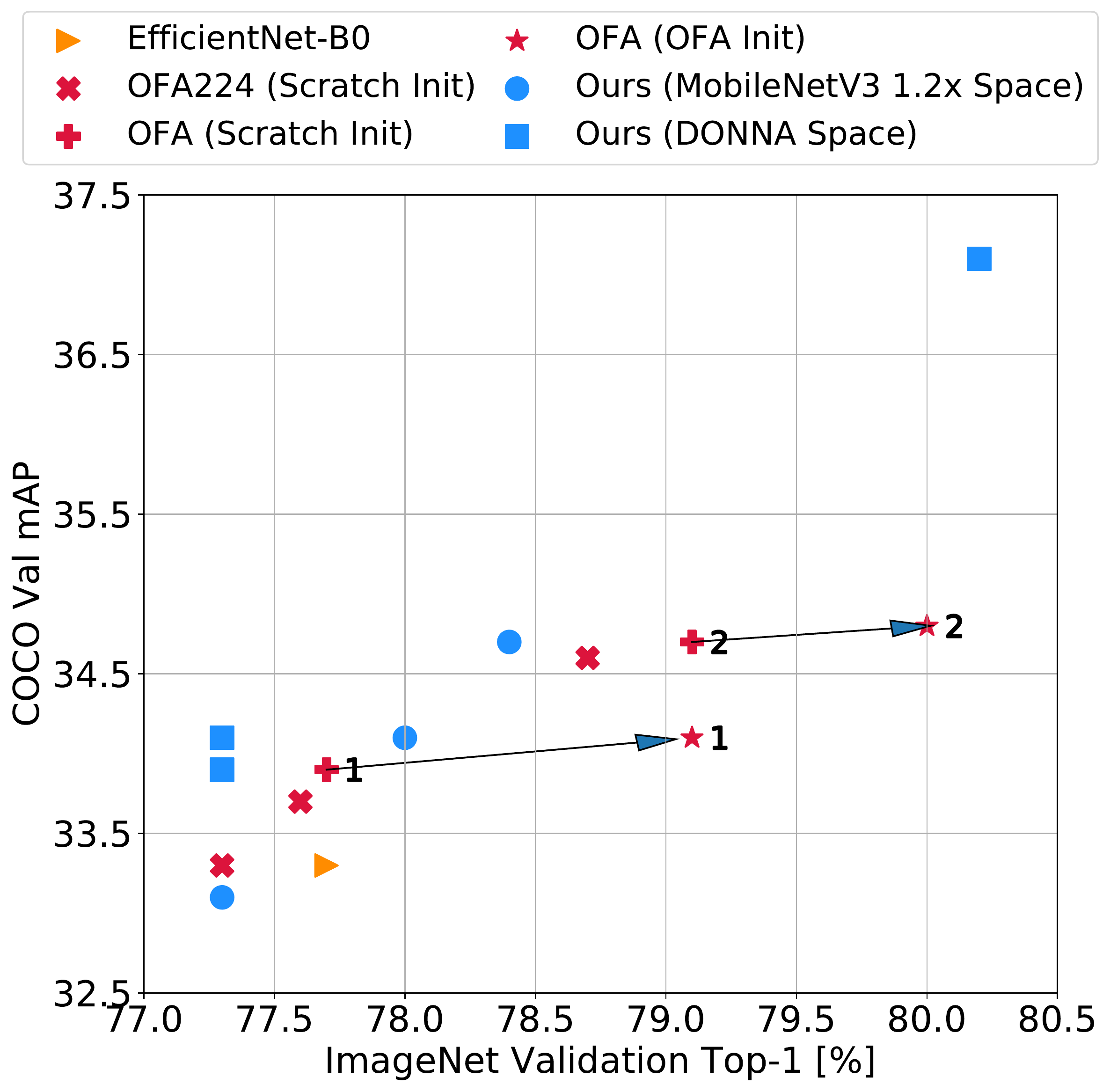}
\caption{Transfer performance of \methodname~backbones to object detection. For \methodname~models, COCO validation mAP correlates well with the ImageNet Validation Top-1 accuracy. This is also the case for OFA models, if they are pretrained on ImageNet under the same or similar circumstances. If the OFA models are trained through progressive shrinking, their higher ImageNet accuracy does not transfer to a higher performance on MS-COCO.}
\label{fig:transfer}
\end{figure}
In this section, we further investigate the transfer properties of \methodname~backbones in an object detection task. Our data hints towards two conclusions: (1) ImageNet top-1 validation is a good predictor for COCO mAP if models are sampled from a similar search space and if they are trained using the same hyperparameters and starting from the same initialization and (2) higher accuracies on ImageNet achieved through progressive shrinking in OFA do not transfer to significantly higher COCO mAP.  The models under study are the same set as in Section \ref{subsec:experiments/detection}.

These conclusions are apparent from Figure~\ref{fig:transfer}. Here, we plot the COCO Val mAPs of the detection architectures against the ImageNet Val top-1 accuracies of their respective backbones.
First, we see that OFA models trained from scratch (OFA Scratch and OFA224) and models found in the similar MobileNetV3 ($1.2\times$) search space through \methodname, transfer very similarly to COCO. Models found in the \methodname~search space reach higher COCO mAP than expected based on their ImageNet top-1 accuracy.  We suspect that such bias occurs because instead of strictly relying on depthwise convolutions, which is the case for MobileNetV3 (1.2$\times$) space, grouped convolutions are used in the \methodname~search space.
Second, we find that while OFA models with OFA training obtain around 1.0-1.5 percent higher accuracy on ImageNet~\cite{imagenet} than the same models trained from scratch, this increased accuracy does not transfer to a meaningful gain in downstream tasks such as object detection. This phenomenon is illustrated in Figure~\ref{fig:transfer}, where the same OFA models are trained on MS-COCO, either starting from weights trained on ImageNet from scratch or starting from weights obtained through progressive shrinking on ImageNet. For one of these models, the $1.4\%$ gain in ImageNet validation accuracy only translates into $0.1\%$ higher mAP on COCO. This observation motivates the choice that throughout the text, we compare to OFA-models which are trained from scratch rather than through progressive shrinking. 


\section{\methodname~for Vision Transformers} 
\label{sec:supplementary/vit}

DONNA can be trivially applied to Vision Transformers~\cite{dosovitskiy2020image}, without any conceptual change to the base algorithm. In this experiment, we use vit-base-patch16-224 from~\cite{ofa_repo} as a teacher model for which we define a related hierarchical search space. Vit-base-patch16-224 is split into 4 DONNA-blocks, each containing 3 ViT blocks (self-attention+MLP) as defined in the original paper~\cite{dosovitskiy2020image}. For every block, we vary the following parameters: 

\begin{itemize}
    \item Vit-block \textit{depth} varies $\in$ \{1,2,3\}
    \item The \textit{embedding dimension} can be scaled down to $50\%$ of the original embedding dimension $\in$\{$50\%$,$75\%$,$100\%$\}, equivalent to $\in$\{384,576,768\} internally in the DONNA-block.
    \item The \textit{number of heads} used in attention varies from 4-to-12 $\in$ \{4,8,12\}.
    \item The \textit{mlp-ratio} can be varied from 2-4 $\in$ \{2,3,4\}. Larger mlp-ratios indicate larger MLP's per block.
\end{itemize}

\begin{figure}[t]
\centering
\includegraphics[width=.99\linewidth]{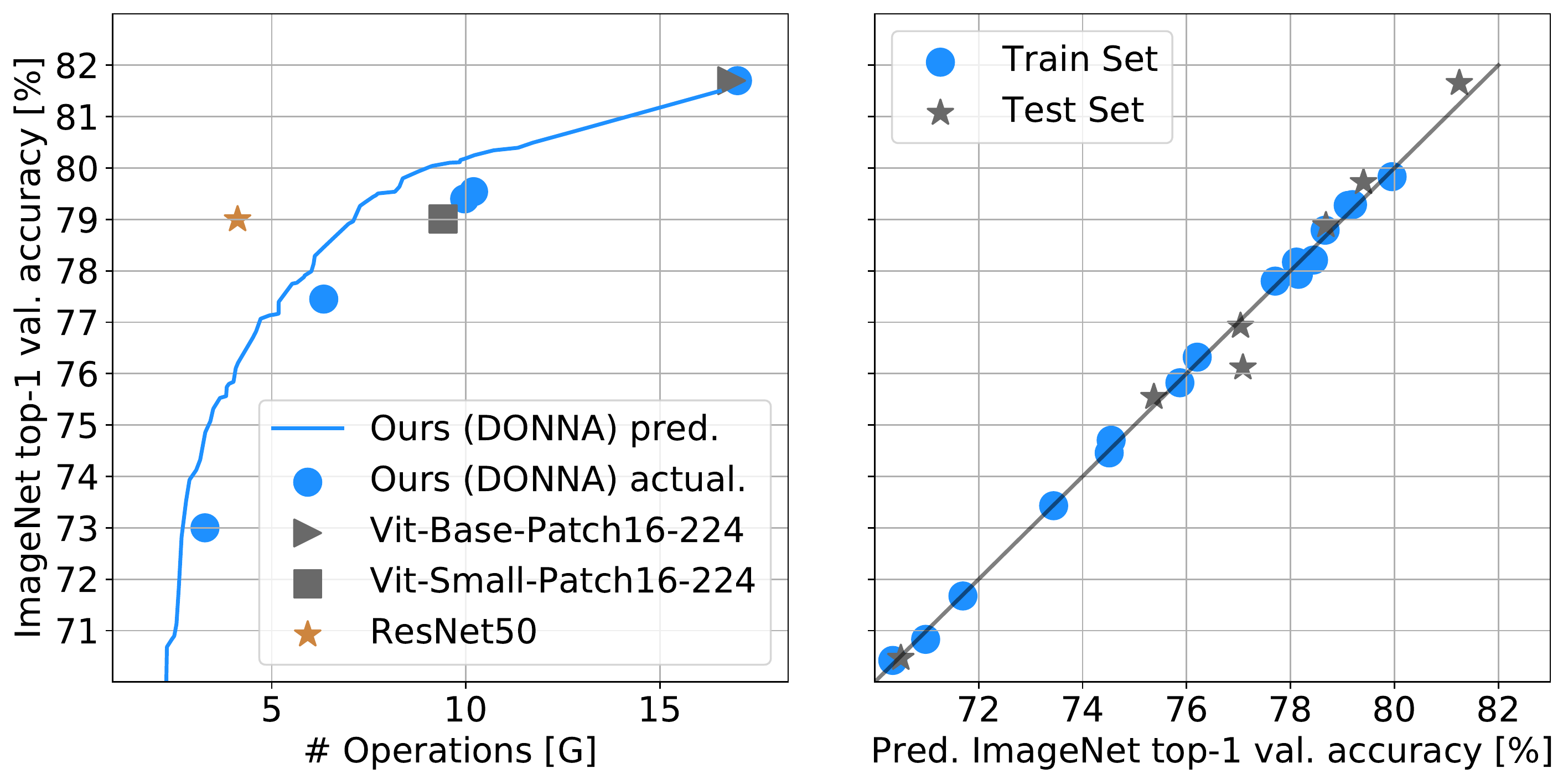}
\caption{(left) Pareto Optimal models found through a DONNA search in a search space based on vit-base-patch16-224 finetuned on ImageNet from ImageNet21k. vit-base-patch16-224 is pretrained on ImageNet21k and finetuned on ImageNet. vit-small-patch16-224 is taken from~\cite{ofa_repo}, and trained using the same pipeline as the DONNA models. (right) Performance of accuracy predictor for the ViT compression case.}
\label{fig:vit}
\end{figure}

Potentially, sequence length can be searched over as well, but this is not done in this example. The \textit{Block Library} is built using the BKD process, requiring $4\times3\times3\times3=135$ epochs of total training to model a fairly small search space of .5M architectures. The \textit{Architecture Library} exists out of 23 uniformly sampled architectures in this search space, finetuned for 50 epochs on ImageNet~\cite{imagenet}, using a large CNN model as a teacher until convergence. The latter process is calibrated such that the original teacher model (vit-base-patch16-224), initialized with weights from the \textit{Block Library} achieves the accuracy of the teacher model after these 50 epochs. Note that our reliance on such finetuning and knowledge distillation allows extracting knowledge without access to full datasets, in this case ImageNet21k. Finally, we use the Block- and Architecture libraries to train an accuracy predictor and execute an evolutionary search targeting minimization of the number of operations. Figure~\ref{fig:vit}(left) illustrates the results of this search, showing that our search in this space allows finding a pareto set of models. In terms of number of operations, this ViT-based search space does not outperform ResNet-50. Figure~\ref{fig:vit}(right) illustrates the quality of the accuracy predictor, on a limited set of ViT architectures.
\section{Search space extension to Quantized Networks} 
\label{sec:supplementary/qnas}

The \methodname~accuracy predictor extends to search spaces different from the one it has been trained for, see Section~\ref{subsubsec:experiments/search-space-design}. This is a major advantage of \methodname~, as it enables us to quickly extend pre-existing NAS results without the need to create an extended Architecture Library and without retraining the accuracy predictor. For details on this, see Section \ref{subsubsec:experiments/search-space-design} and Fig. \ref{fig:linear_predictor} for a discussion on this using ShiftNets~\cite{shiftnet}. This section illustrates that the \methodname~accuracy predictor is not only portable across layer types, but also across different compute precisions, i.e. when using quantized INT8 operators.

To demonstrate this, let us consider the MobileNetV3 (1.2$\times$) search space.
First, we build and train a \methodname~accuracy predictor for full-precision (FP) networks and then test this predictor for networks with weights and activations quantized to 8 bits (INT8).
The search space includes k $\in \{3,5,7\}$; expand $\in \{3,4,6\}$; depth $\in \{2,3,4\}$; activation $\in\{ReLU/Swish\}$; attention $\in \{None/SE\}$; and channel-scaling $\in \{0.5\times, 1.0\times\}$. We build a complete Block Library in FP; sampling 43 FP networks as an Architecture Library and finetuning them to collect the training data for the FP accuracy predictor model. 
Second, we quantize the Block Library using the Data-Free-Quantization (DFQ)~\cite{dfq} post training quantization method using 8 bits weights and activations (INT8). The quantized Block Library now provides the quality metrics for quantized blocks, which can be used as inputs to the FP accuracy predictor to predict INT8 accuracy. 
Finally, we test the FP accuracy predictor model on a test set of INT8 networks. For this, we sample 20 networks whose INT8-block quality is within the range of the train set of the accuracy predictor.
These networks are first finetuned in FP using the procedure outlined in section~\ref{sec:search_method} and then quantized to INT8 using DFQ~\cite{dfq}.


\begin{figure}[t]
\centering
\includegraphics[width=.99\linewidth]{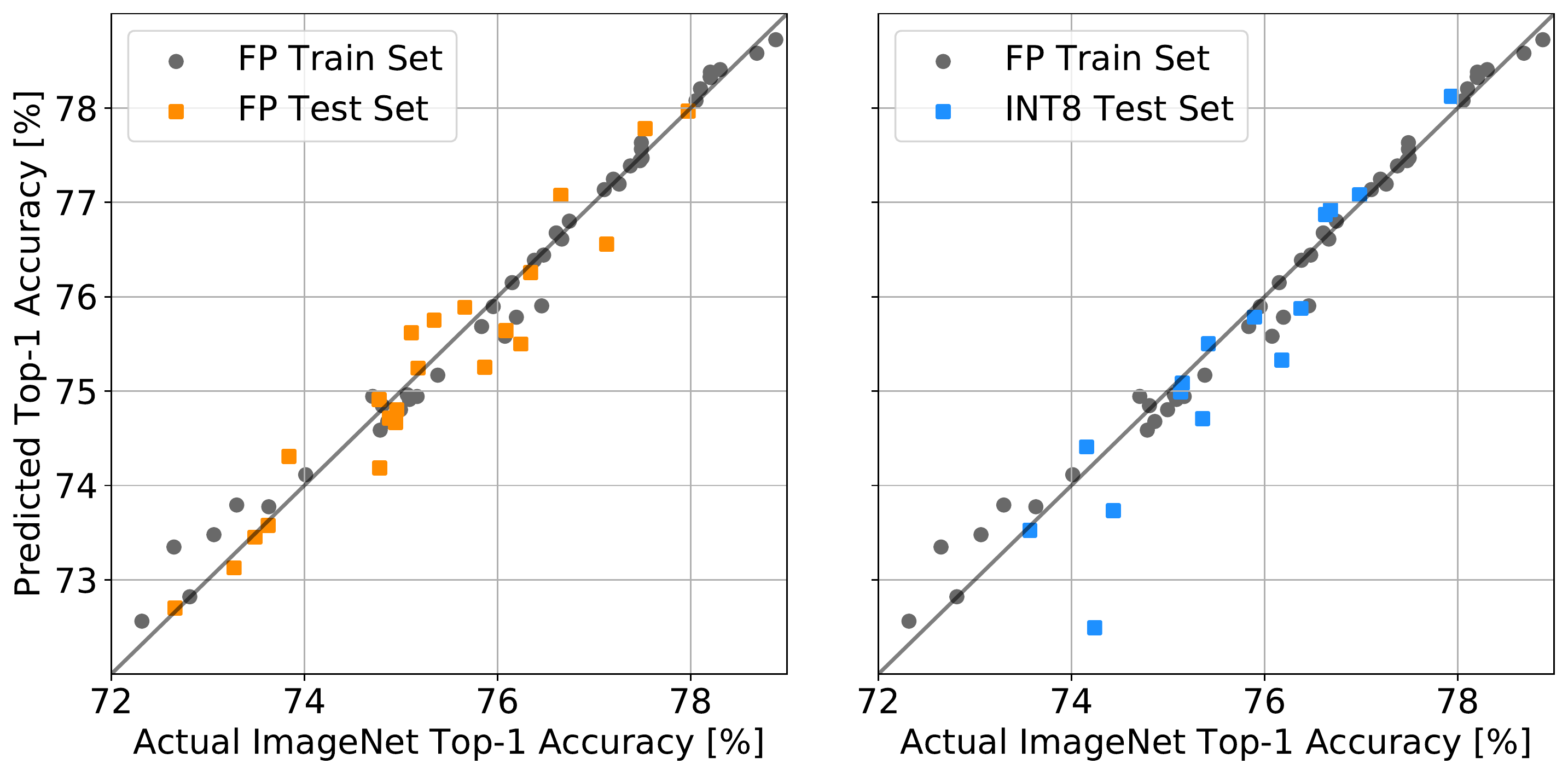}
\caption{Validation of the accuracy prediction model trained on FP networks and tested on FP networks (right) and INT8 networks (left). Kendal-Tau values are 0.85, and 0.86 respectively for the Test-FP and Test-INT8 sets.}
\label{fig:supplementary/qnasvalid}
\end{figure}

Figure \ref{fig:supplementary/qnasvalid} illustrates the FP predictor can be used to directly predict the performance of INT8 networks, indicating that DONNA search spaces can indeed be trivially extended to include INT8 precision. Fig.~\ref{fig:supplementary/qnasvalid}(left) shows FP train and test data for the accuracy predictor model. Fig.~\ref{fig:supplementary/qnasvalid}(right) shows FP train and INT8 test data using the same FP accuracy predictor. 
Formally, we compare the performance of this predictor on the FP and INT8 test set by comparing the achieved prediction MSE and Kendal-Tau (KT)~\cite{kendall1938new}. We can observe that there are no outliers when using the predictor to predict the accuracy of INT8 networks. MSE for the FP test set is 0.13 and 0.34 for the INT8 test set. MSE for INT8 is higher because of the noise introduced by the quantization process. Nonetheless the KT-ranking is 0.85 for FP test set and 0.86 for the INT8 test set demonstrating that the accuracy predictor can be used for INT8-quantized models.

\section{Comments on random search} 
\label{sec:supplementary/random-search}

DONNA clearly outperforms random search. In random search, networks are sampled randomly with some latency or complexity constraint and trained from scratch. This can be very costly if the accuracy of these architectures varies widely, as is the case in a large and diverse search space. On top of that, any expensive random search would have to be repeated for every target accuracy or latency on any new hardware platform. This is in stark contrast with DONNA, where the accuracy predictor is reused for any target accuracy, latency and hardware platform.

\begin{figure}[t]
\centering
\includegraphics[width=.80\linewidth]{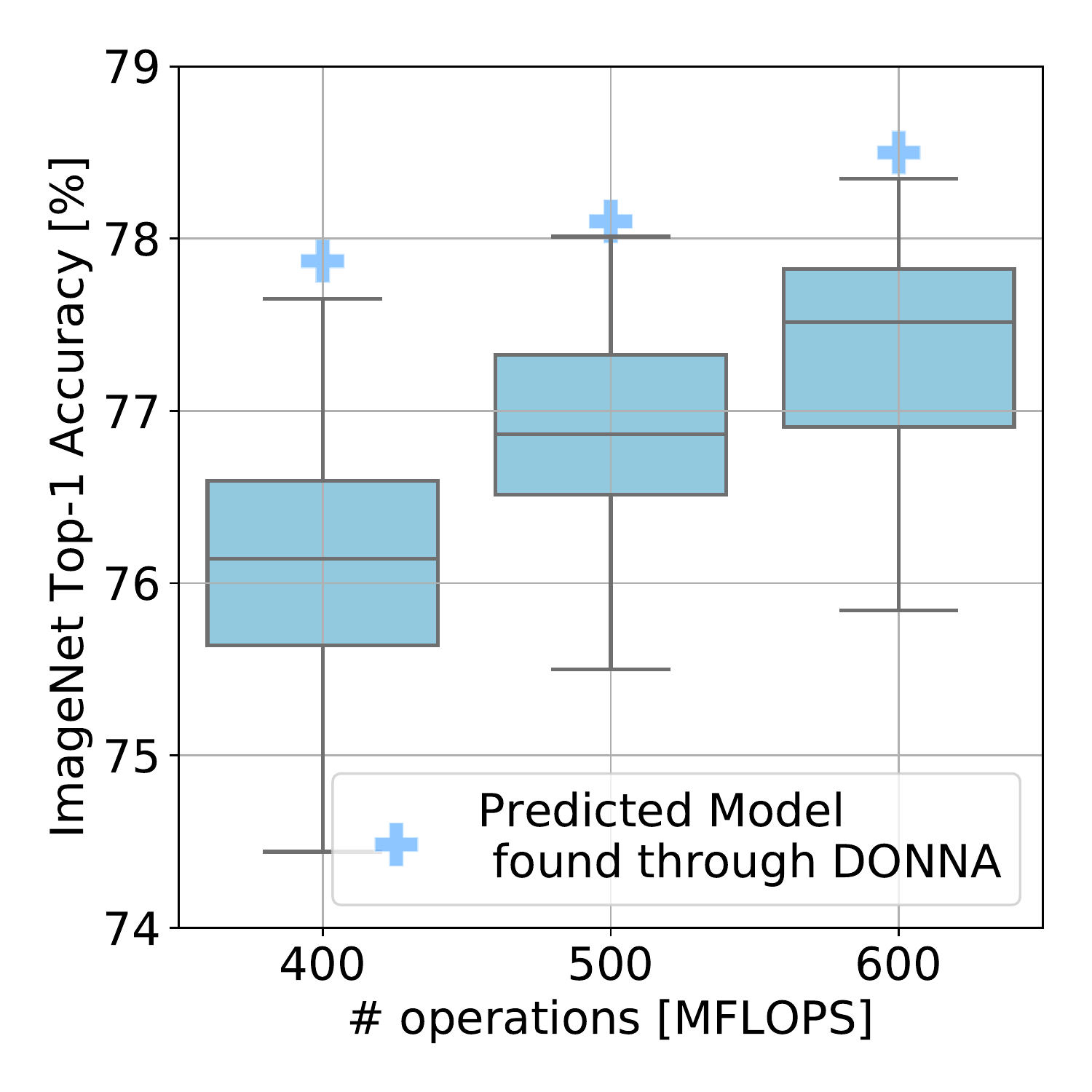}
\caption{Comparing statistics of random architectures in the MobileNetV3 (1.2$\times$) search-space, as predicted by the DONNA accuracy predictor, to the predicted accuracy of models found through DONNA at the same number of operations.}
\label{fig:supplementary/random_search}
\end{figure}

Fig.~\ref{fig:supplementary/random_search} illustrates box-plots for the predicted accuracy on ImageNet-224 for networks randomly sampled in the MobileNetV3 (1.2$\times$) search space, at 400 +/-5 (190 samples), 500 +/- 5 (77 samples) and 600 +/- 5 (19 samples) million operations (MFLOPS). The box shows the quartiles of the dataset while the whiskers extend to show the rest of the distribution. According to the accuracy predictor, randomly sampled architectures at 400M operations are normally distributed with a mean and standard deviation of 76.2$\%$ and 0.7$\%$ respectively. Based on this, only around 2$\%$ of the randomly sampled architectures will have an accuracy exceeding 77.6$\%$. So, when performing true random search for the 400M operation target, training 100 architectures for 450 epochs (45000 epochs in total) will likely yield 2 networks exceeding 77.6$\%$. In contrast, after building the accuracy predictor for MobileNetV3 (1.2$\times$) in 1500 epochs, DONNA finds an architecture achieving 77.5$\%$ at 400M operations in just 50 epochs, see Figure~\ref{fig:supplementary/search-space-overview}(mid-left). This is close to a 900$\times$ advantage if the start up cost is ignored, a reasonable assumption at a large amount of targets. In summary, the total cost of random search scales as $N\times450\times\#$latency-targets$\times\#$platforms, where $N$ is the number of trained samples for every latency-target on every platform. DONNA scales as $50\times\#$latency-targets$\times\#$platforms when many latency-targets and hardware platforms are being considered, meaning the initial costs of building the reusable accuracy predictor can be ignored. 

Predictor-based random search could also be used as a replacement for the NSGA-II evolutionary search algorithm \cite{nsga_ii} in DONNA. However, NSGA-II is known to be more sample efficient than random search in a multi-objective setting~\cite{hughes2006multi}. This is also illustrated in Figure~\ref{fig:supplementary/random_search}, where NSGA-II finds networks with a higher predicted accuracy than random search, given the 190 (400M), 77 (500M) and 19 (600M) samples for every target. In this NSGA-II, a total of 2500 samples was generated and measured during the search, covering the full search-space ranging from 150-800M operations.


\section{Model Visualizations}
\label{subsec:supplementary/model-visualizations}

\begin{figure*}[!t]
\centering
\begin{subfigure}{.8\textwidth}
  \centering
  \includegraphics[width=.8\linewidth]{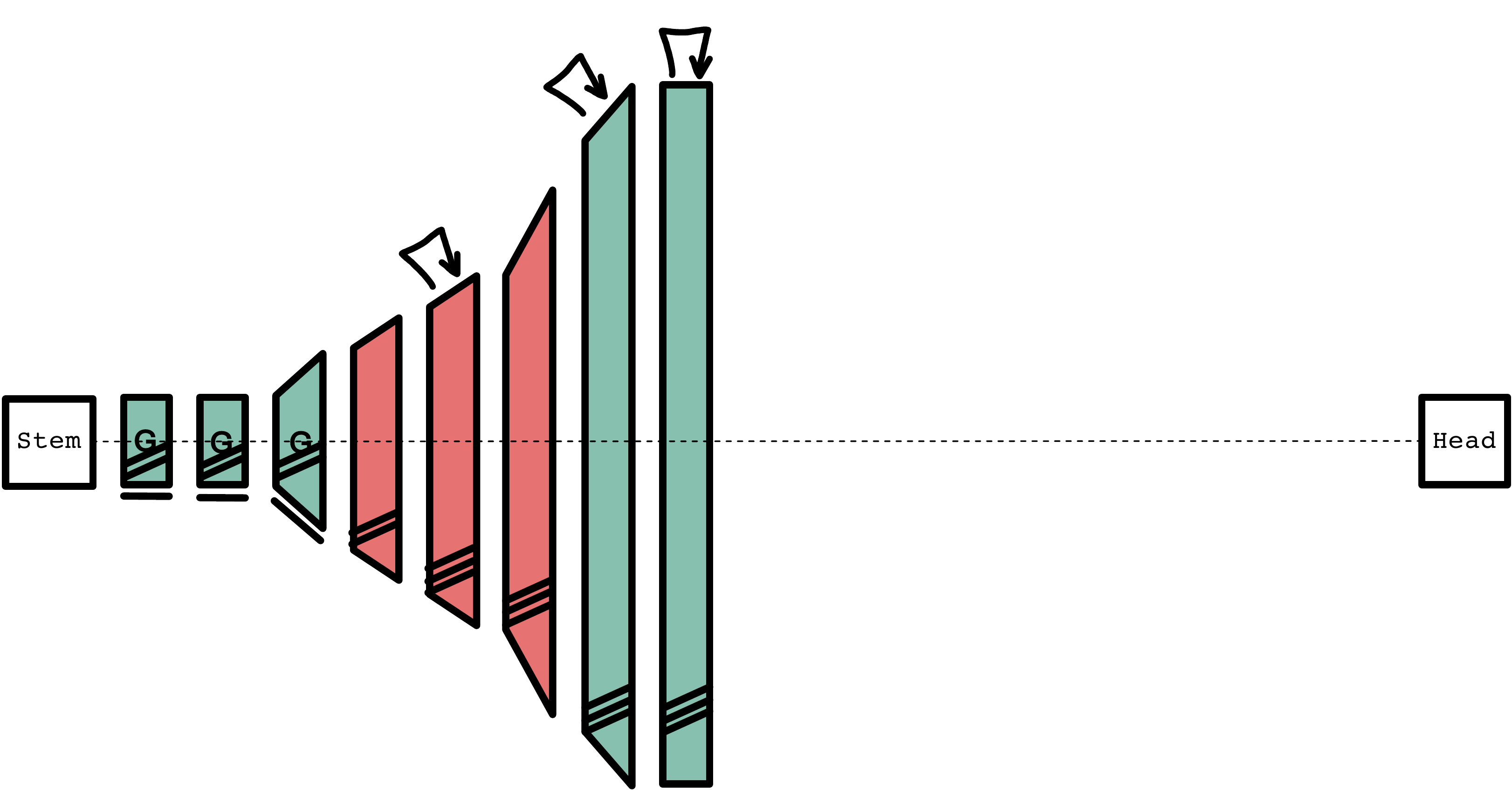}
  \caption{\methodname~for a simulator targeting tensor compute in a mobile SoC, $73.7\%$ at $0.45\times$ the latency of EfficientNet-B0.}
  \label{fig:params}
\end{subfigure}
\begin{subfigure}{.8\textwidth}
  \centering
  \includegraphics[width=.8\linewidth]{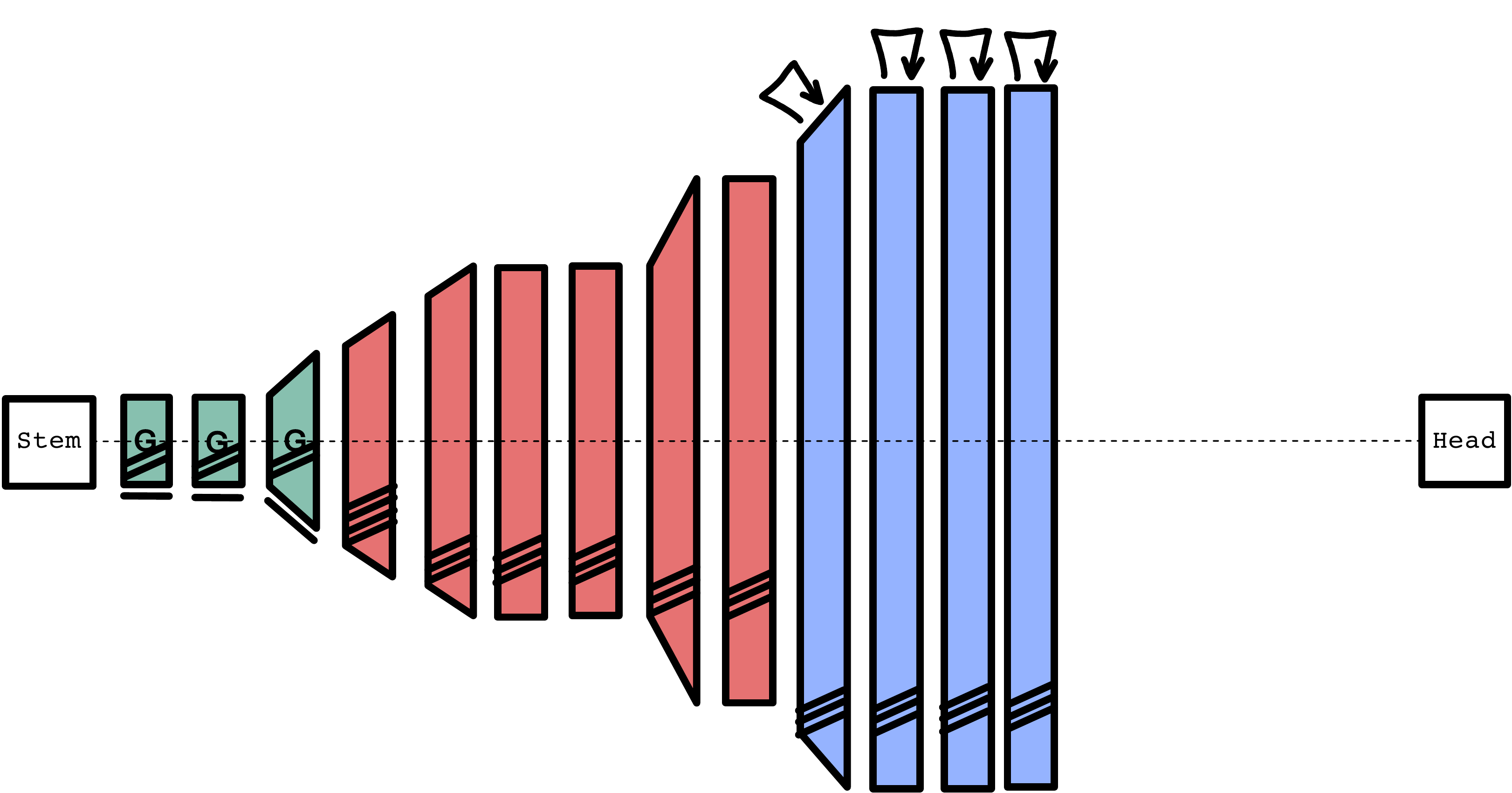}
  \caption{\methodname~for a simulator targeting tensor compute in a mobile SoC, $77.25\%$ at $0.60\times$ the latency of EfficientNet-B0.}
  \label{fig:flops}
\end{subfigure}
\begin{subfigure}{.8\textwidth}
  \centering
  \includegraphics[width=.8\linewidth]{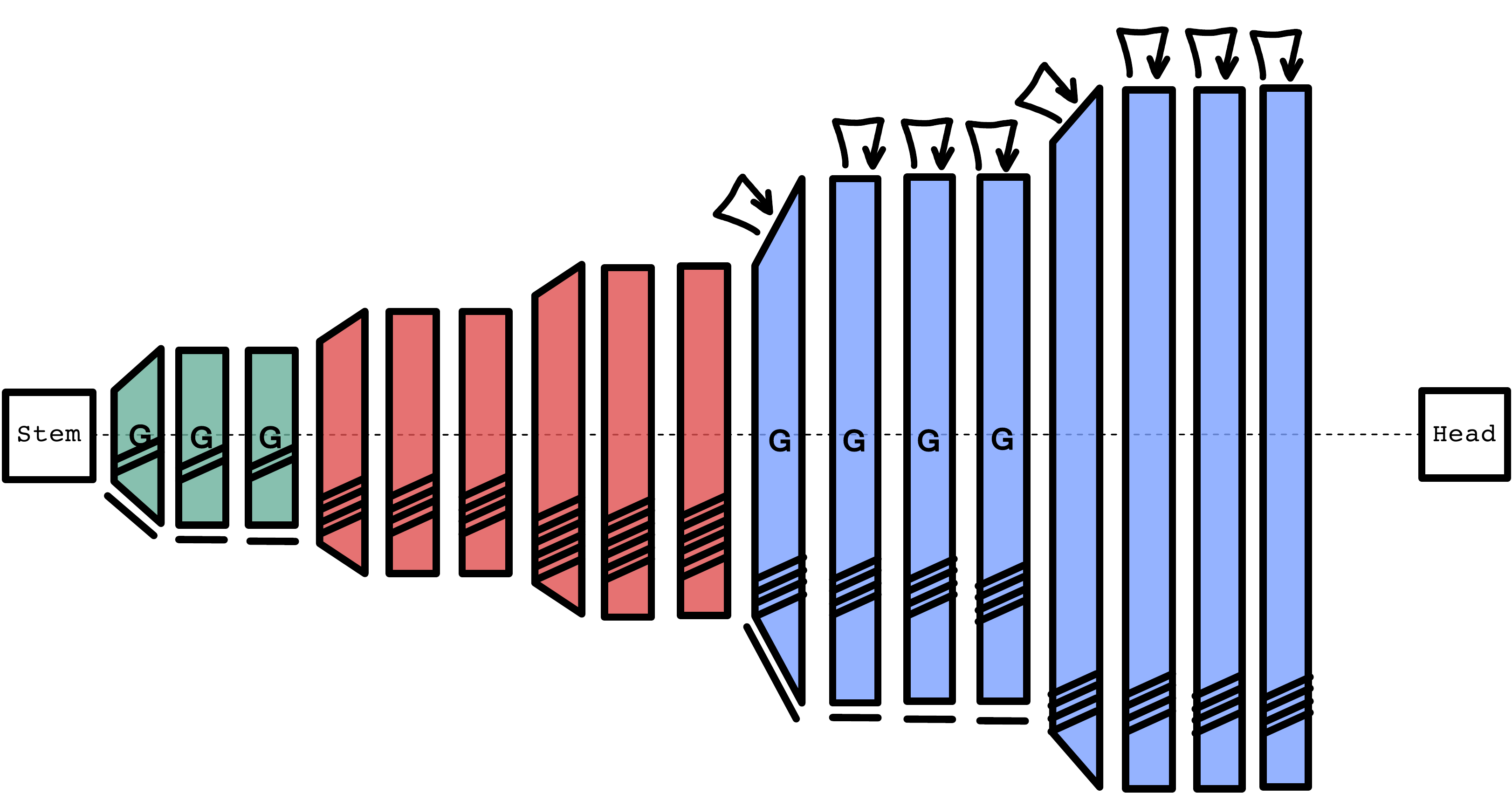}
  \caption{\methodname~for a simulator targeting tensor compute in a mobile SoC, $80.2\%$ at $1.25\times$ the latency of EfficientNet-B0.}
  \label{fig:gpu}
\end{subfigure}%
\caption{Example models found through ~\methodname~in the~\methodname~search space, Pareto-optimal on ImageNet for a simulator targeting tensor compute units in a mobile SoC. The Box-color indicates kernel-size: green (3), blue (5) and red (7). Every box is an inverted residual bottleneck with depthwise-separable (plain) or grouped (line under the box) layers. The box height is related to the number of channels. The number of dashed lines per box indicate the expansion rate, the arrows on top indicate whether or not Squeeze-and-Excite and Swish is used. }
\label{fig:magellan_visualizations}
\end{figure*}

\begin{figure*}[!t]
\centering
\begin{subfigure}{.8\textwidth}
  \centering
  \includegraphics[width=.8\linewidth]{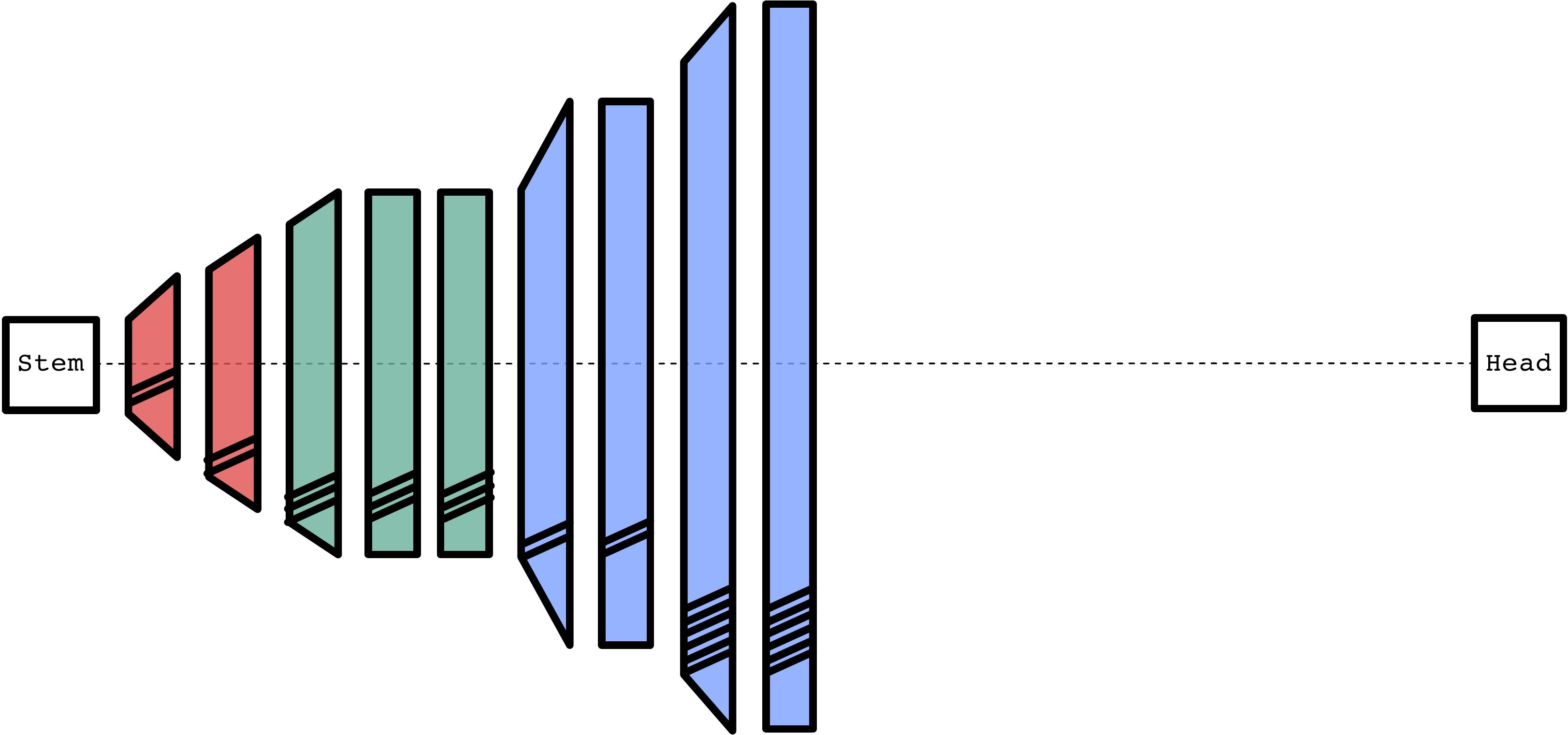}
  \caption{\methodname~for Nvidia V100, batch-size 32, $73\%$ top-1 @7.7ms on ImageNet.}
  \label{fig:params}
\end{subfigure}
\begin{subfigure}{.8\textwidth}
  \centering
  \includegraphics[width=.8\linewidth]{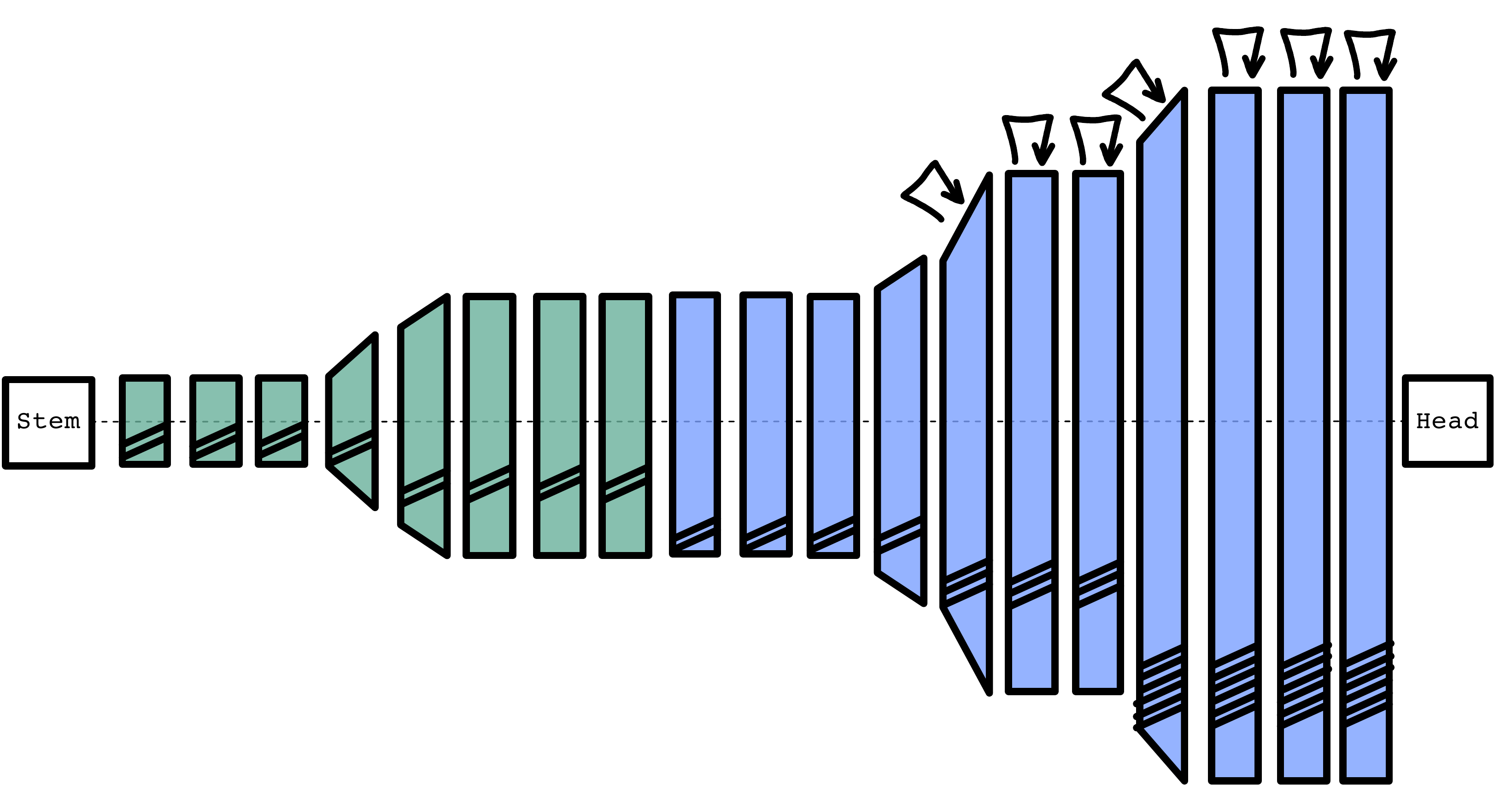}
  \caption{\methodname~for Nvidia V100, batch-size 32, $75.7\%$ top-1 @10.15ms on ImageNet.}
  \label{fig:flops}
\end{subfigure}
\begin{subfigure}{.8\textwidth}
  \centering
  \includegraphics[width=.8\linewidth]{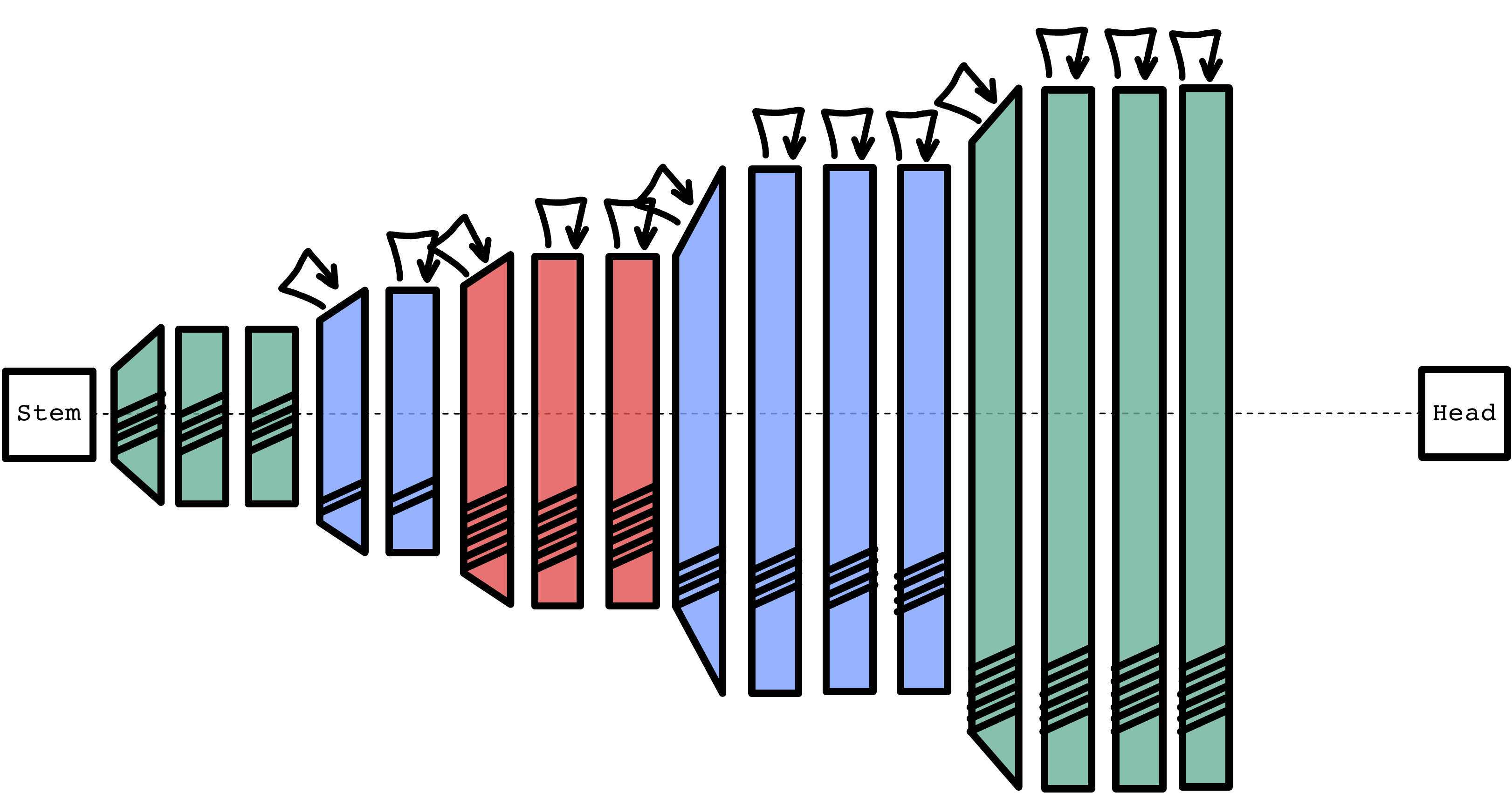}
  \caption{\methodname~for Nvidia V100, batch-size 32, $79.5\%$ top-1 @24.9ms on ImageNet.}
  \label{fig:gpu}
\end{subfigure}%
\caption{Example models found through ~\methodname~in the~\methodname~search space, Pareto-optimal on ImageNet for the Nvidia V100 GPU, with a Batch-Size of 32. The Box-color indicates kernel-size: green (3), blue (5) and red (7). Every box is an inverted residual bottleneck with depthwise-separable (plain) or grouped (line under the box) layers. The box height is related to the number of channels. The number of dashed lines per box indicate the expansion rate, the arrows on top indicate whether or not Squeeze-and-Excite and Swish is used. }
\label{fig:gpu_visualizations}
\end{figure*}

\begin{figure*}[!t]
\centering
\begin{subfigure}{.8\textwidth}
  \centering
  \includegraphics[width=.8\linewidth]{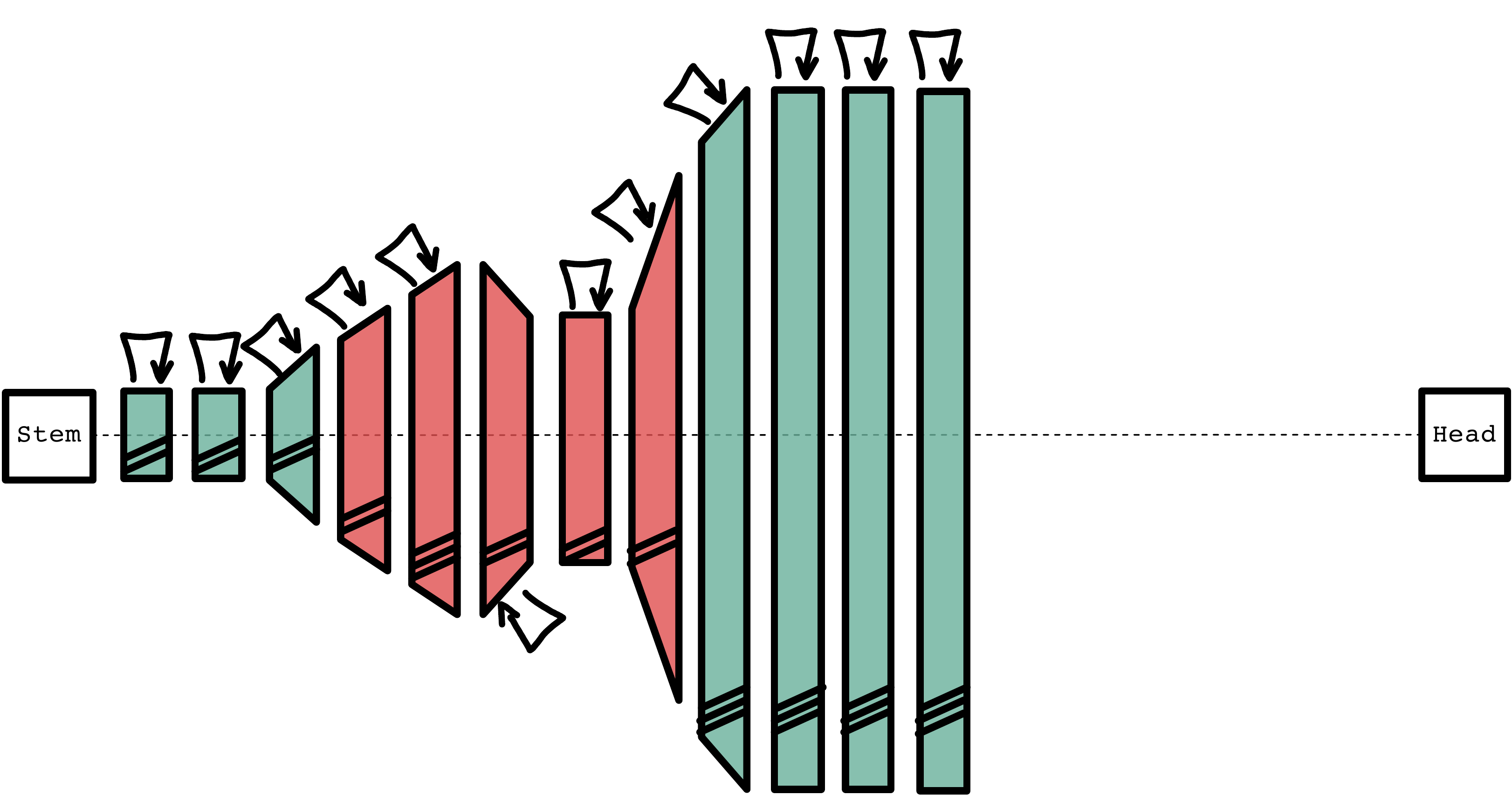}
  \caption{\methodname~for number of operations, 75.3$\%$ @ 331 MFLOP}
  \label{fig:params}
\end{subfigure}
\begin{subfigure}{.8\textwidth}
  \centering
  \includegraphics[width=.8\linewidth]{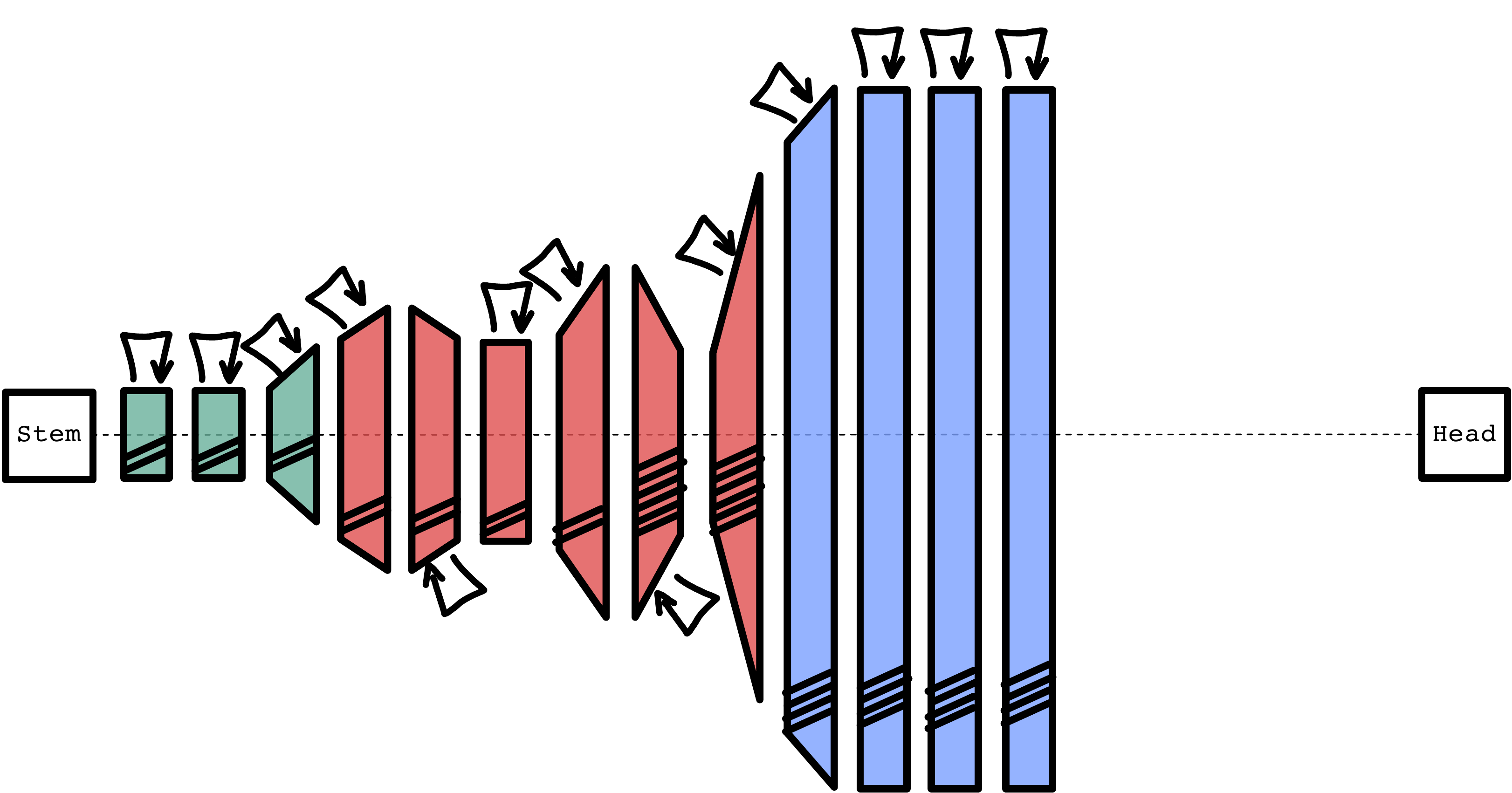}
  \caption{\methodname~for number of operations, 76.6$\%$ @ 405 MFLOP}
  \label{fig:flops}
\end{subfigure}
\begin{subfigure}{.8\textwidth}
  \centering
  \includegraphics[width=.8\linewidth]{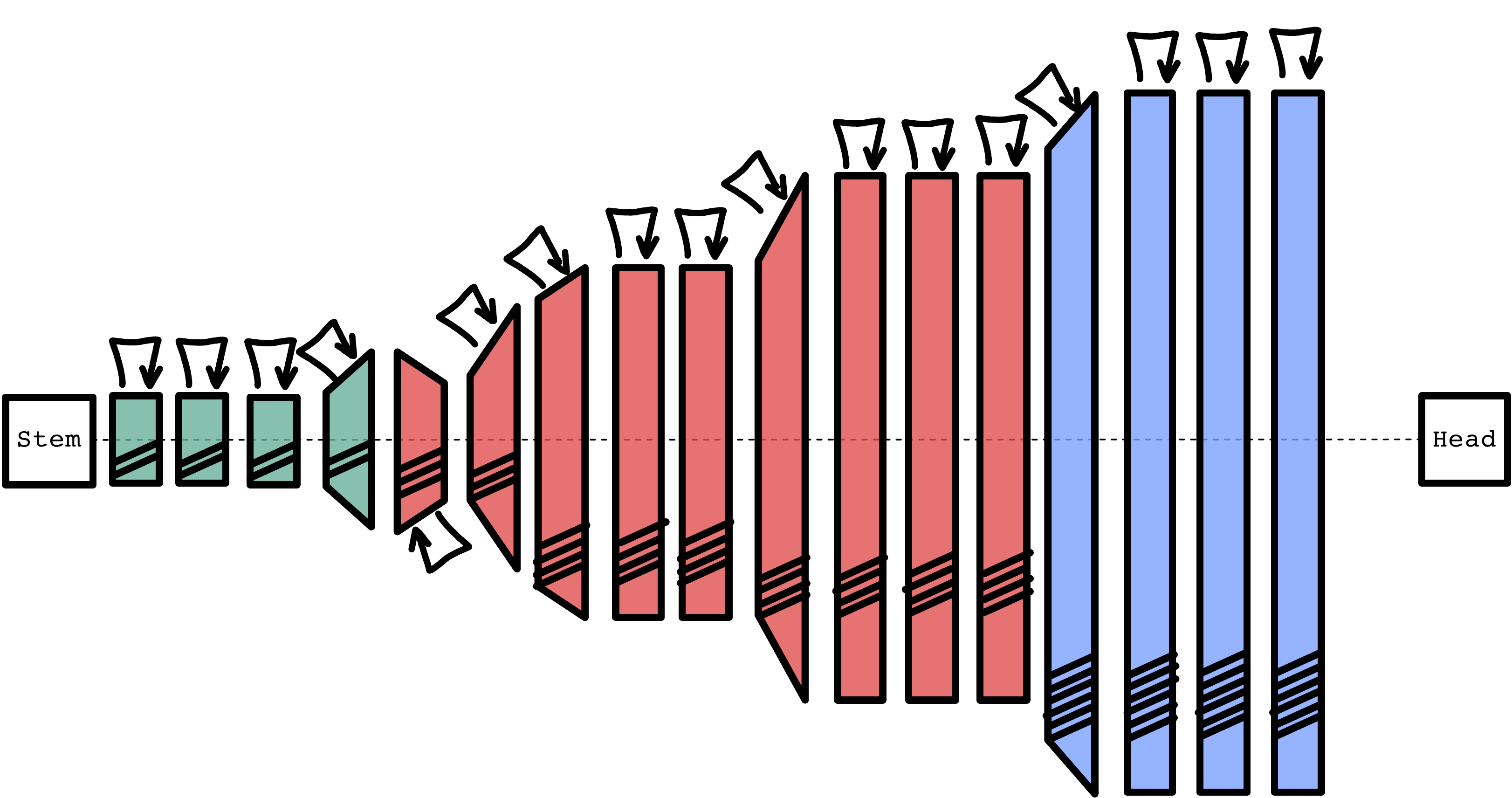}
  \caption{\methodname~for number of operations, 79.1$\%$ @ 800 MFLOP}
  \label{fig:gpu}
\end{subfigure}%
\caption{Example models found through ~\methodname~in the~\methodname~search space, Pareto-optimal on ImageNet optimized for the number of operations. The Box-color indicates kernel-size: green (3), blue (5) and red (7). Every box is an inverted residual bottleneck with depthwise-separable (plain) or grouped (line under the box) layers. The box height is related to the number of channels. The number of dashed lines per box indicate the expansion rate, the arrows on top indicate whether or not Squeeze-and-Excite is used. Note that it is optimal to use SE in every block.}
\label{fig:flops_visualization}
\end{figure*}

\begin{figure*}[!t]
\centering
\begin{subfigure}{.8\textwidth}
  \centering
  \includegraphics[width=.8\linewidth]{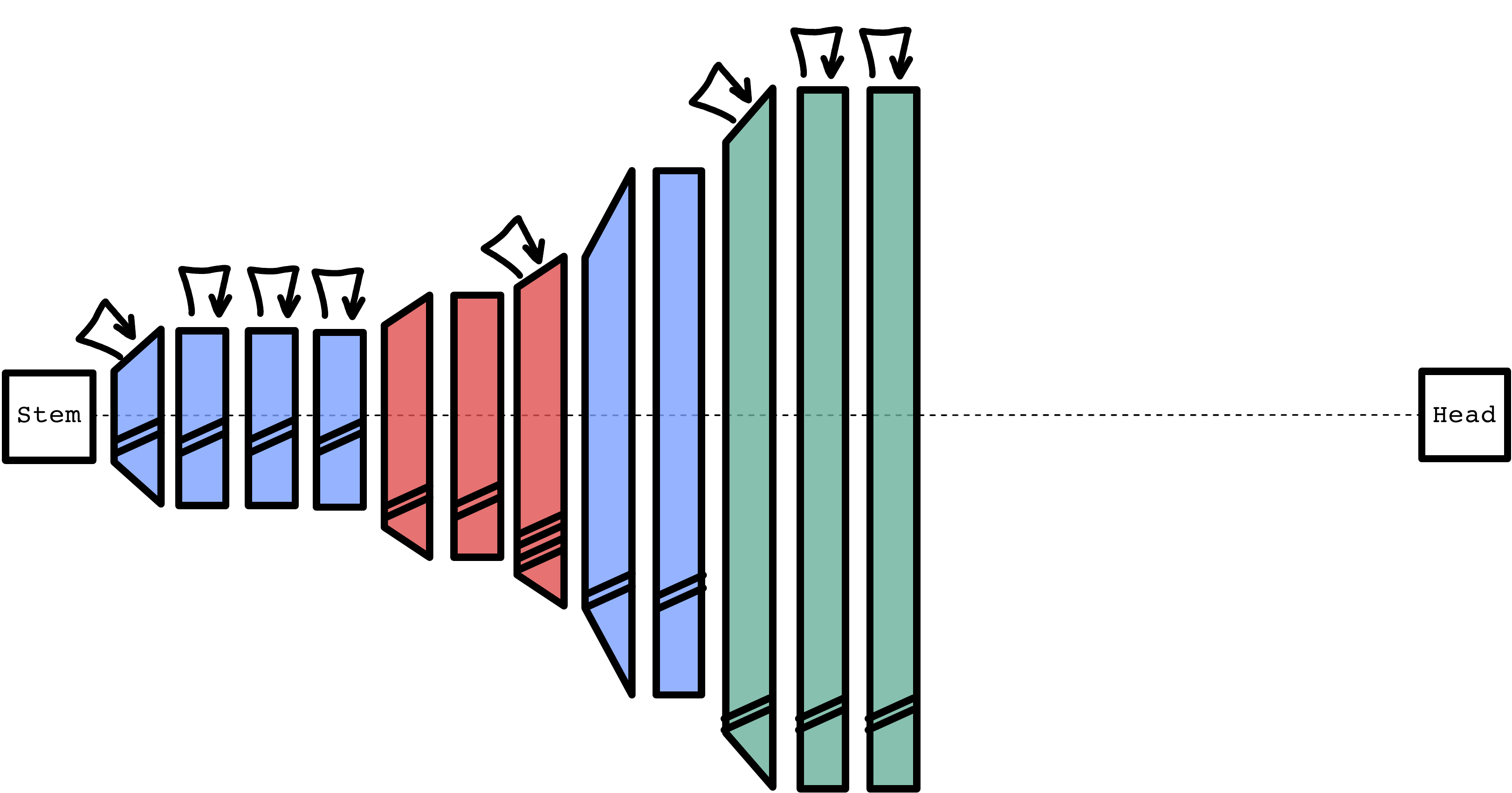}
  \caption{\methodname~for number of parameters, 76.0$\%$ @ 3.5 million parameters. }
  \label{fig:params}
\end{subfigure}
\begin{subfigure}{.8\textwidth}
  \centering
  \includegraphics[width=.8\linewidth]{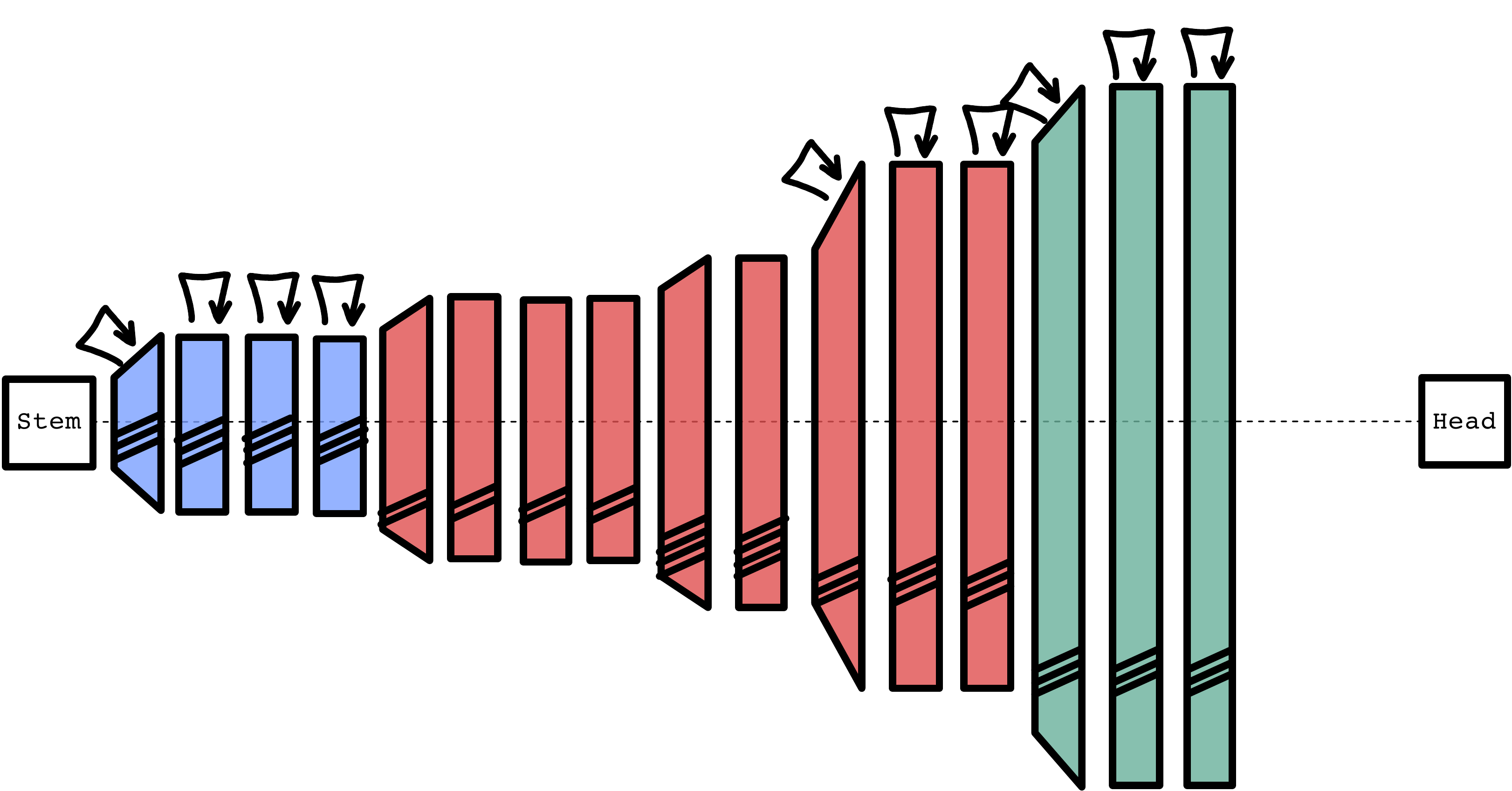}
  \caption{\methodname~for number of parameters, 78.3$\%$ @ 5.16 million parameters.}
  \label{fig:flops}
\end{subfigure}
\begin{subfigure}{.8\textwidth}
  \centering
  \includegraphics[width=.8\linewidth]{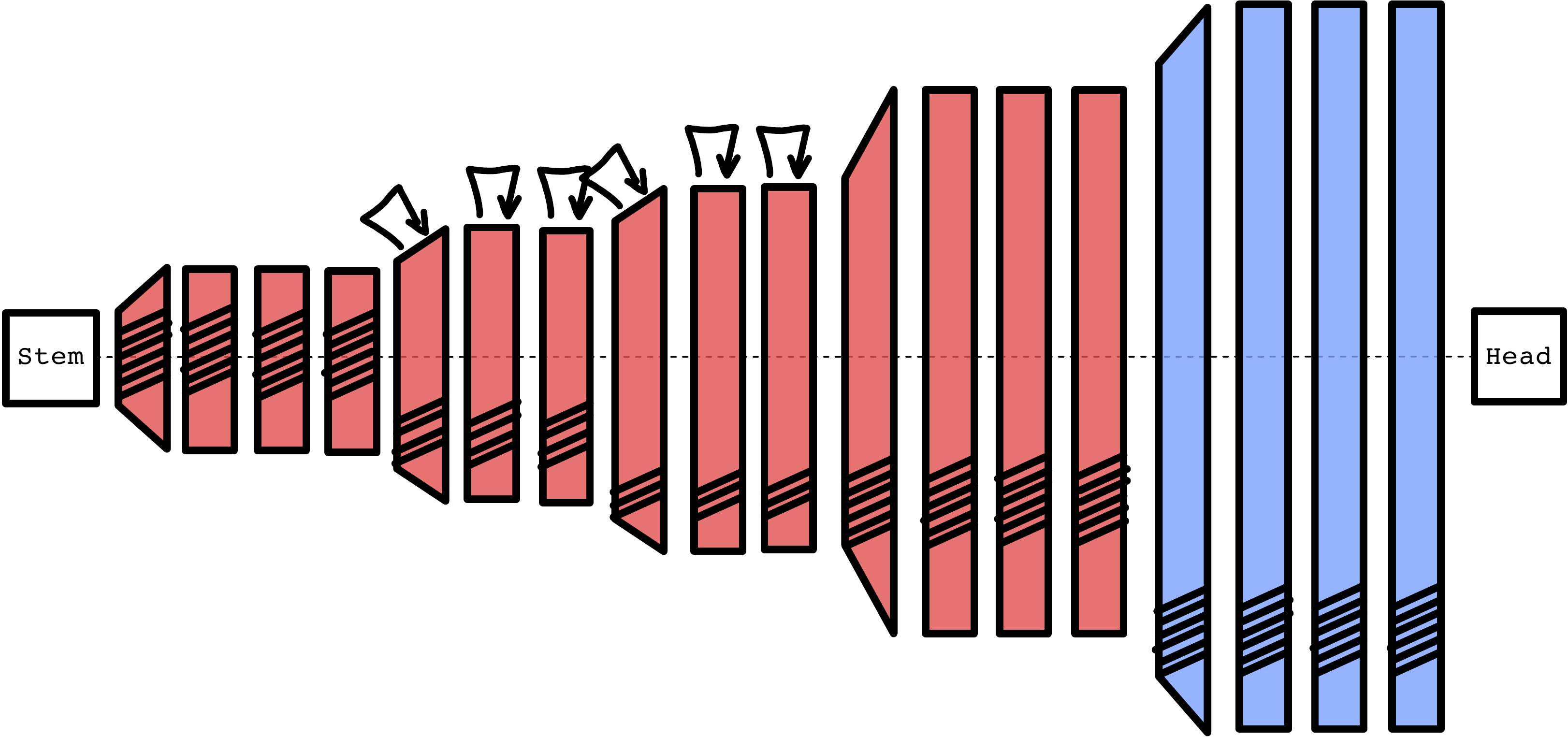}
  \caption{\methodname~for number of parameters, 80.0$\%$ @ 7.5 million parameters.}
  \label{fig:gpu}
\end{subfigure}%
\caption{Example models found through ~\methodname~in the~\methodname~search space, Pareto-optimal on ImageNet optimized for the number of parameters. The Box-color indicates kernel-size: green (3), blue (5) and red (7). Every box is an inverted residual bottleneck with depthwise-separable (plain) or grouped (line under the box) layers. The box height is related to the number of channels. The number of dashed lines per box indicate the expansion rate, the arrows on top indicate whether or not Squeeze-and-Excite and Swish is used.}
\label{fig:params_visualization}
\end{figure*}

\begin{figure*}[!t]
\centering
\begin{subfigure}{.8\textwidth}
  \centering
  \includegraphics[width=.8\linewidth]{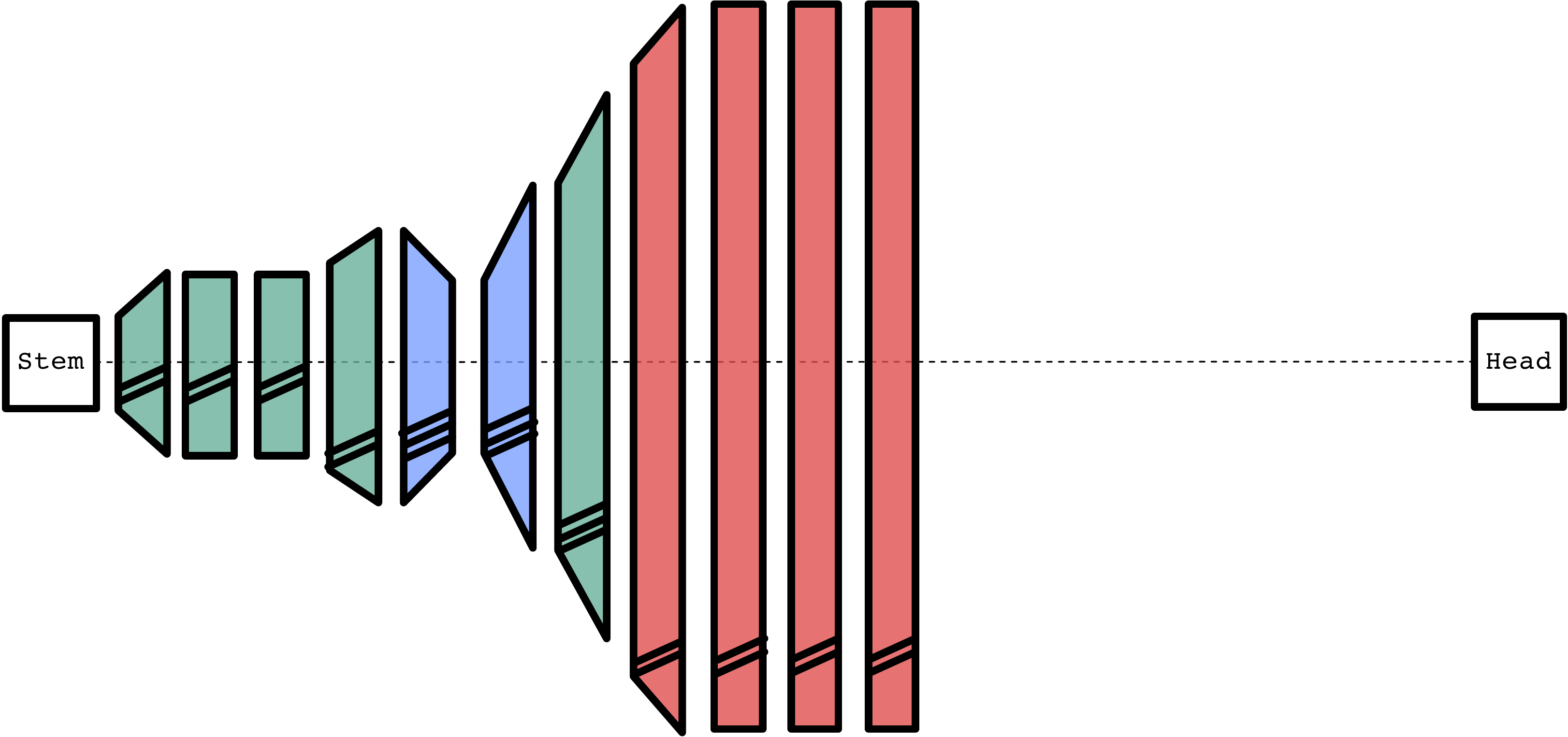}
  \caption{\methodname~for the Samsung S20 GPU, 72.7$\%$ @ 6.5 ms.}
  \label{fig:params}
\end{subfigure}
\begin{subfigure}{.8\textwidth}
  \centering
  \includegraphics[width=.8\linewidth]{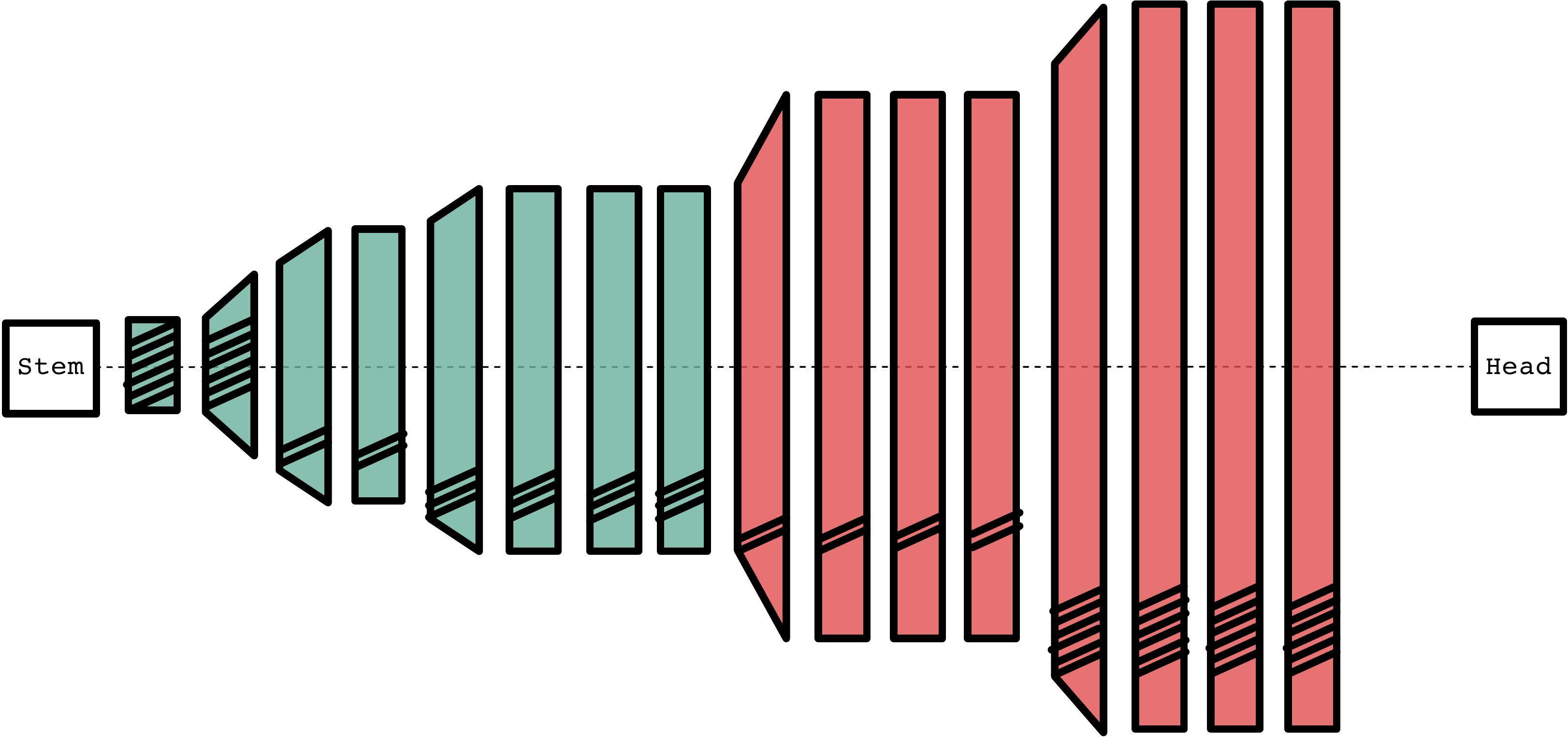}
  \caption{\methodname~for the Samsung S20 GPU, 76.96$\%$ @ 10.5 ms.}
  \label{fig:flops}
\end{subfigure}
\begin{subfigure}{.8\textwidth}
  \centering
  \includegraphics[width=.8\linewidth]{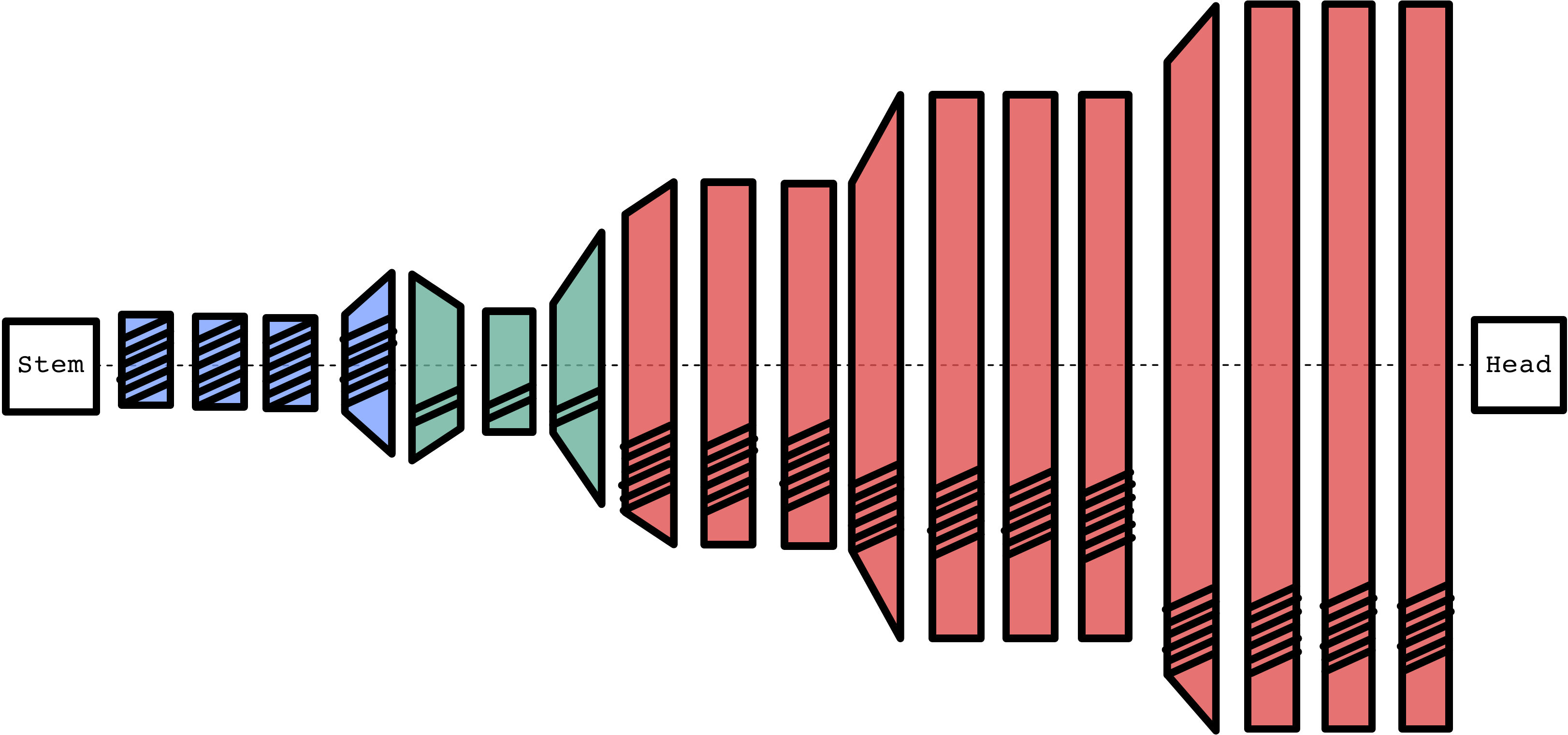}
  \caption{\methodname~for the Samsung S20 GPU, 78.9$\%$ @ 16.2 ms.}
  \label{fig:gpu}
\end{subfigure}%
\caption{Example models found through ~\methodname~in the~\methodname~search space, Pareto-optimal on ImageNet optimized for the Samsung S20 GPU. The Box-color indicates kernel-size: green (3), blue (5) and red (7). Every box is an inverted residual bottleneck with depthwise-separable (plain) or grouped (line under the box) layers. The box height is related to the number of channels. The number of dashed lines per box indicate the expansion rate, the arrows on top indicate whether or not Squeeze-and-Excite and Swish is used.}
\label{fig:kona_visualization}
\end{figure*}

Figures~\ref{fig:magellan_visualizations},~\ref{fig:gpu_visualizations},~\ref{fig:flops_visualization},~\ref{fig:params_visualization} and~\ref{fig:kona_visualization} visualize some of the diverse network architectures found through~\methodname~in the \methodname~search space. Results are shown for a simulator, the Nvidia V100 GPU, the number of operations, the number of parameters, and the Samsung S20 GPU. Note that all of these networks have different patterns of Squeeze-and-Excite (SE~\cite{squeeze_excite}) and activation functions (whenever SE is used, Swish is also used), channel scaling, expansion rates, and kernel factors, as well as varying network depths. In Figure~\ref{fig:magellan_visualizations}, grouped convolutions are also used as parts of optimal networks as a replacement of depthwise separable kernels. 

\begin{figure*}[!t]
\centering
\begin{subfigure}{.8\textwidth}
  \centering
  \includegraphics[width=.8\linewidth]{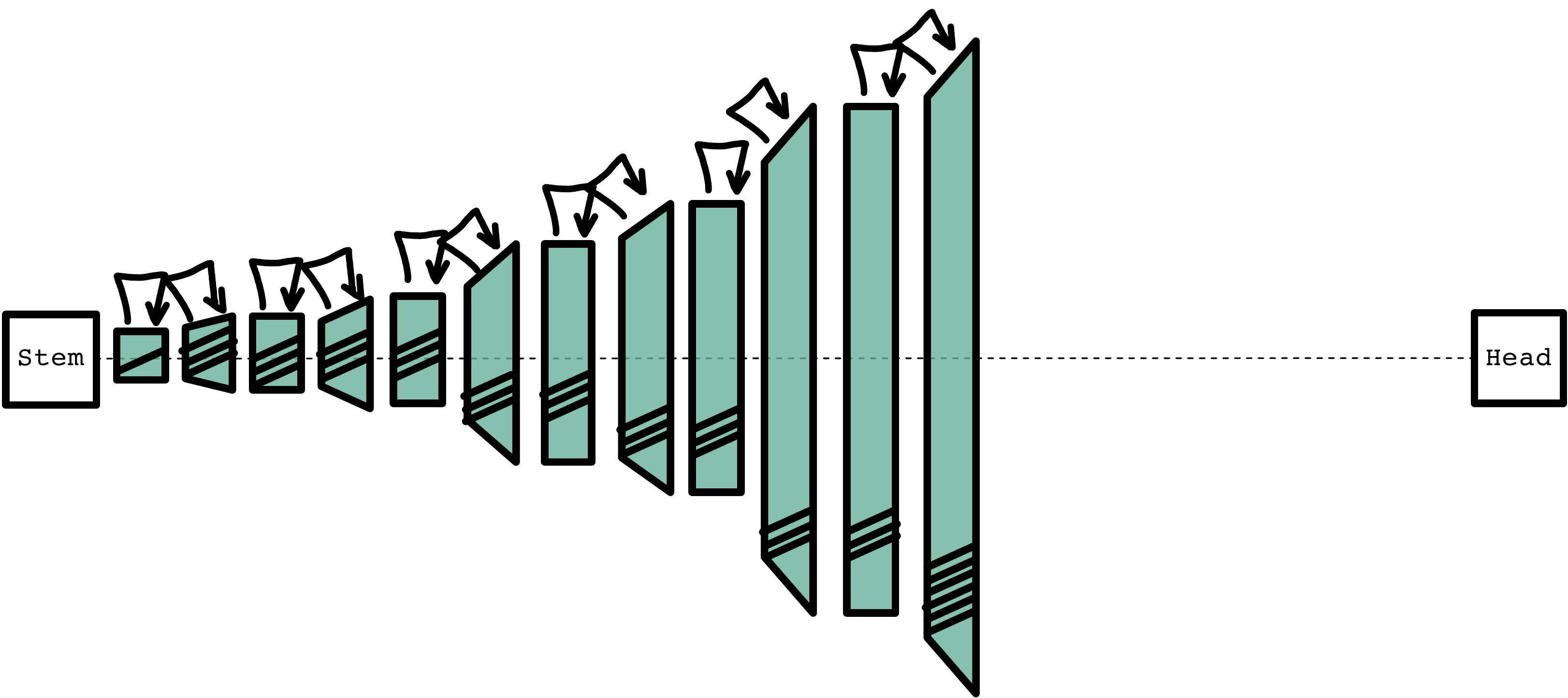}
  \caption{\methodname~for number of operations, 73.4$\%$ @ 185 MFLOP}
  \label{fig:params}
\end{subfigure}
\begin{subfigure}{.8\textwidth}
  \centering
  \includegraphics[width=.8\linewidth]{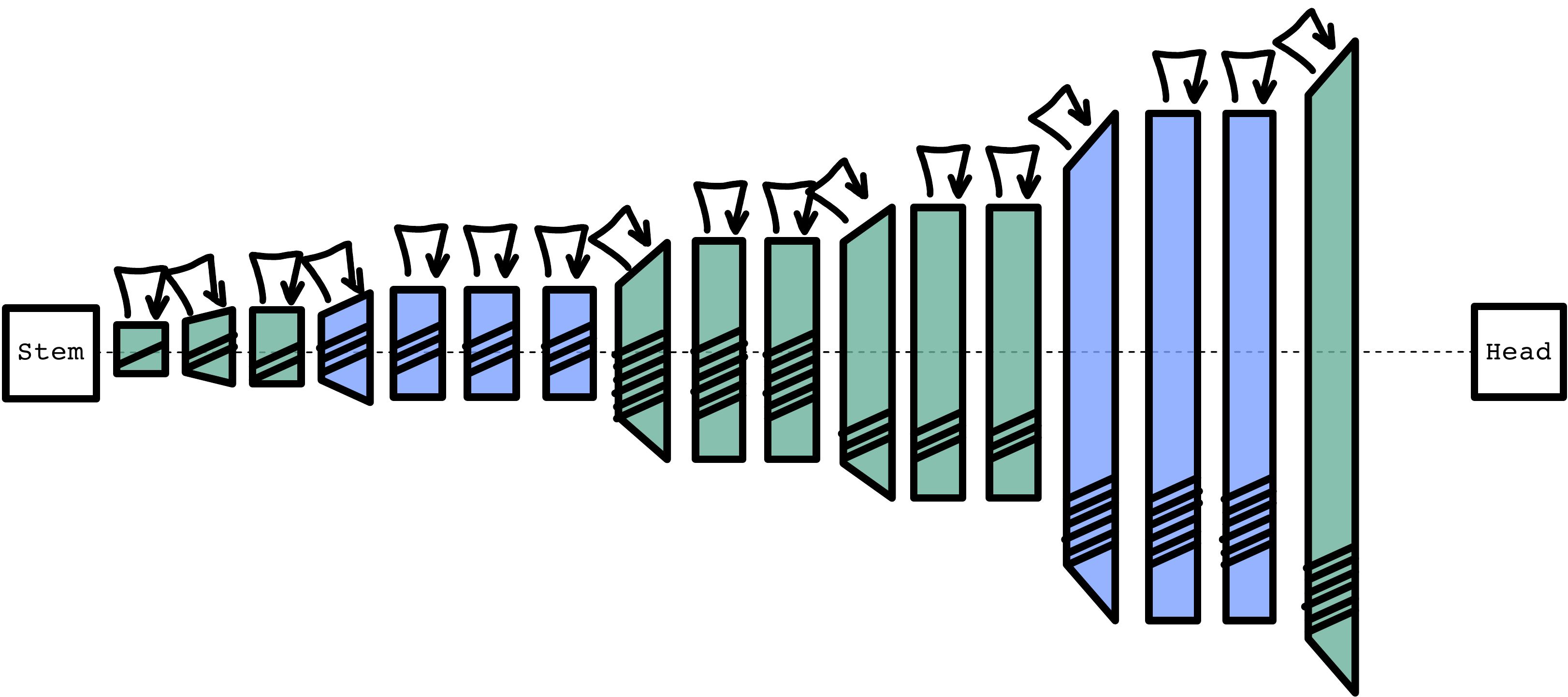}
  \caption{\methodname~for number of operations, 76.6$\%$ @ 301 MFLOP}
  \label{fig:flops}
\end{subfigure}
\begin{subfigure}{.8\textwidth}
  \centering
  \includegraphics[width=.8\linewidth]{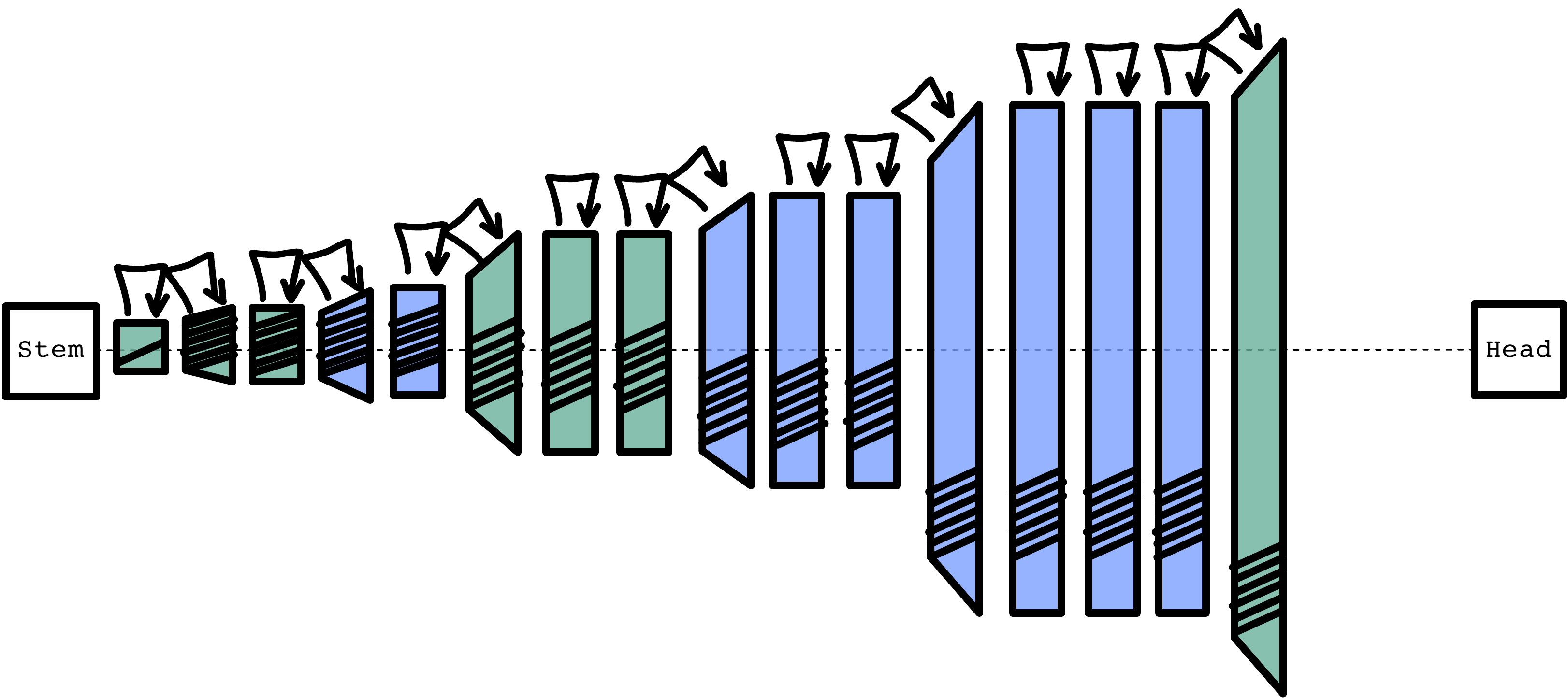}
  \caption{\methodname~for number of operations, 77.7$\%$ @ 405 MFLOP}
  \label{fig:gpu}
\end{subfigure}%
\caption{Example models found through ~\methodname~in the EfficientNet-B0~search space, Pareto-optimal on ImageNet optimized for the number of operations. The Box-color indicates kernel-size: green (3), blue (5) and red (7). Every box is an inverted residual bottleneck with depthwise-separable layers. The box height is related to the number of channels. The number of dashed lines per box indicate the expansion rate, the arrows on top indicate whether or not Squeeze-and-Excite is used. Note that it is optimal to use SE in every block.}
\label{fig:effnet_flops_visualization}
\end{figure*}

\begin{figure*}[!t]
\centering
\begin{subfigure}{.8\textwidth}
  \centering
  \includegraphics[width=.8\linewidth]{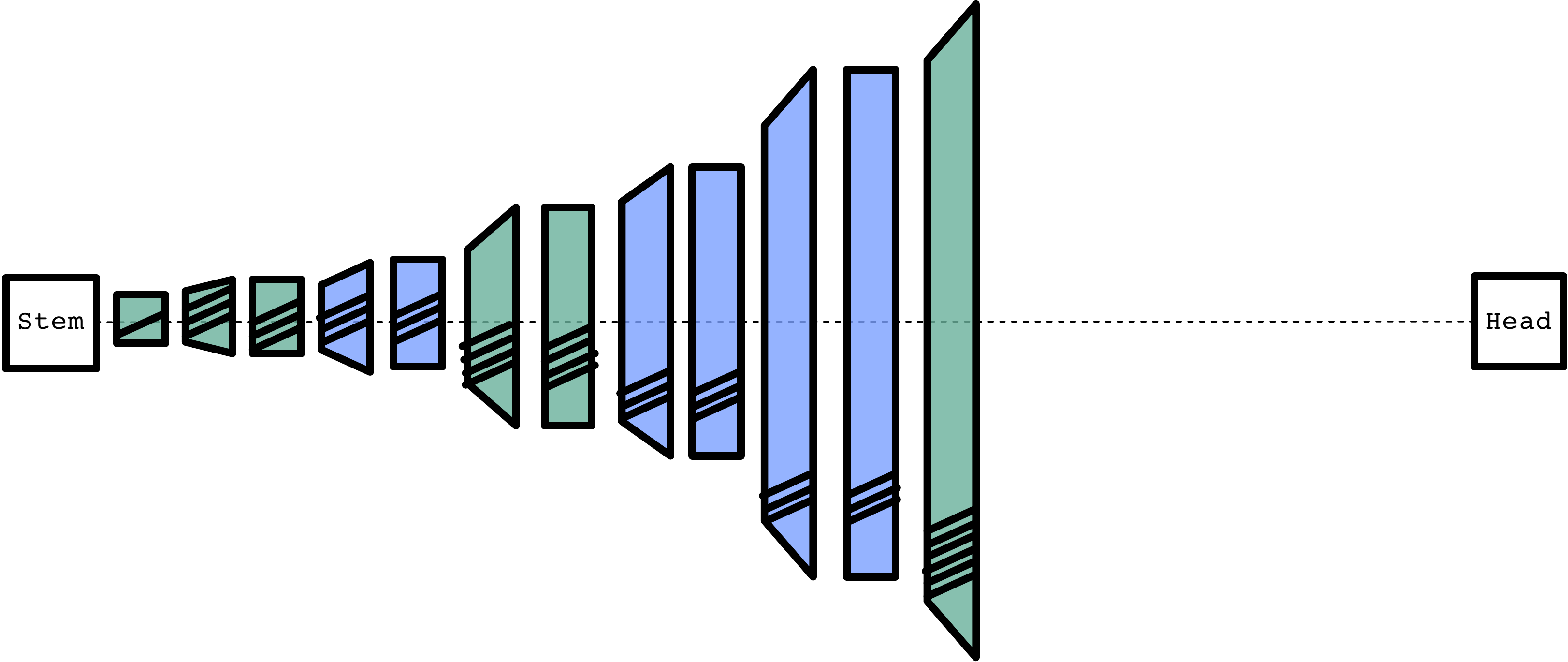}
  \caption{\methodname~for the Samsung S20 GPU, 71.5$\%$ @ 5.2 ms.}
  \label{fig:params}
\end{subfigure}
\begin{subfigure}{.8\textwidth}
  \centering
  \includegraphics[width=.8\linewidth]{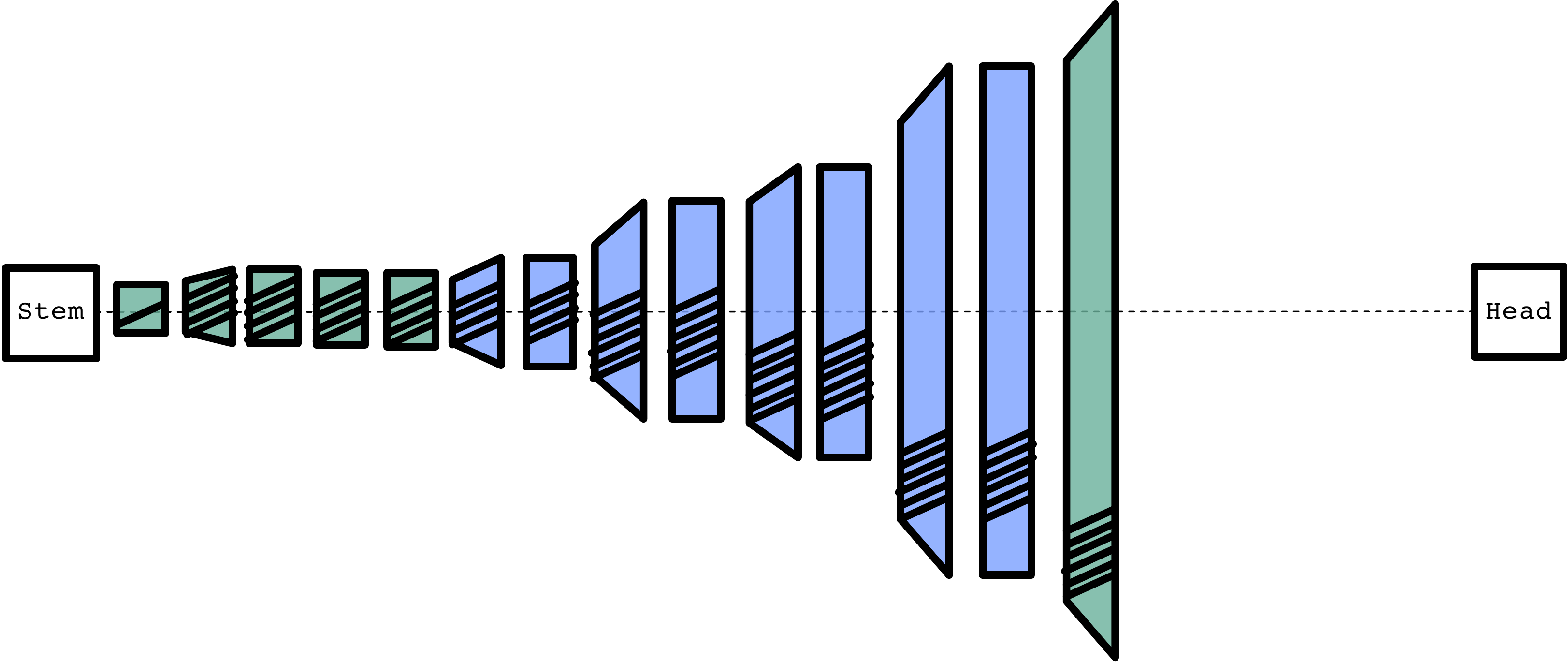}
  \caption{\methodname~for the Samsung S20 GPU, 73.6$\%$ @ 5.95 ms.}
  \label{fig:flops}
\end{subfigure}
\begin{subfigure}{.8\textwidth}
  \centering
  \includegraphics[width=.8\linewidth]{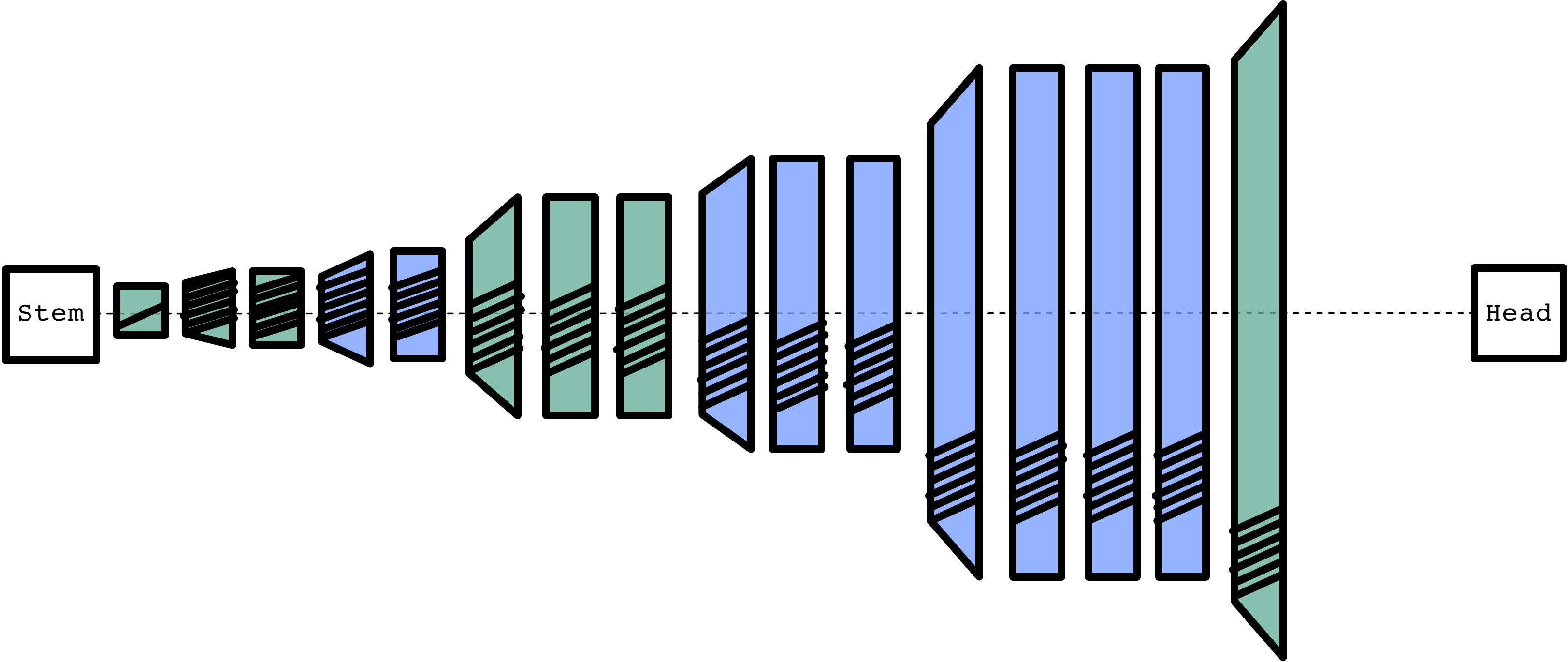}
  \caption{\methodname~for the Samsung S20 GPU, 75.4$\%$ @ 8.8 ms.}
  \label{fig:gpu}
\end{subfigure}%
\caption{Example models found through ~\methodname~in the EfficientNet~search space, Pareto-optimal on ImageNet optimized for the Samsung S20 GPU. The Box-color indicates kernel-size: green (3), blue (5) and red (7). Every box is an inverted residual bottleneck with depthwise-separable layers. The box height is related to the number of channels. The number of dashed lines per box indicate the expansion rate, the arrows on top indicate whether or not Squeeze-and-Excite and Swish is used.}
\label{fig:effnet_kona_visualization}
\end{figure*}

Figure~\ref{fig:effnet_flops_visualization} and~\ref{fig:effnet_kona_visualization} illustrate optimal EfficientNet-Style networks for the number of operations and the Samsung S20 respectively, as taken from Figure~\ref{fig:efficientnet}. Note how these networks are typically narrower, with higher expansion rates than the~\methodname~models, which makes them faster or more efficient in some cases. However, EfficientNet-Style models cannot achieve higher accuracy than $77.7\%$ top-1 on ImageNet validation using $224\times224$ images, while the~\methodname~search space can achieve an accuracy higher than $80\%$ in that case.

\end{document}